\documentclass[journal]{IEEEtran}

\usepackage{graphicx,amssymb,amsmath,amsfonts}

\usepackage[ruled,vlined]{algorithm2e}
\usepackage{array}
\usepackage[caption=false,font=normalsize,labelfont=footnotesize,textfont=footnotesize]{subfig}
\usepackage{textcomp}
\usepackage{stfloats}
\usepackage{url}
\usepackage{verbatim}
\usepackage{cite}
\usepackage{tikz}
\usetikzlibrary{shapes}
\usetikzlibrary{plotmarks}
\usetikzlibrary{shadings}
\usepackage{tablefootnote}

\usepackage{xcolor}
\usepackage{siunitx}
\usepackage{tabularx} 
\usepackage{multirow}
\usepackage{balance}
\usepackage{hyperref}
\usepackage{epstopdf}
\DeclareMathOperator*{\argmin}{arg\,min}

\begin{document}

\title{A Resource-Efficient Decentralized Sequential Planner for Spatiotemporal Wildfire Mitigation}
\author{Josy John,~\IEEEmembership{Student Member,~IEEE}, Shridhar Velhal, Suresh Sundaram,~\IEEEmembership{Senior Member,~IEEE}
\thanks{Josy John, Shridhar Velhal, and Suresh Sundaram are with the Department of Aerospace Engineering, Indian Institute of Science, Bengaluru, India. \href{mailto:josyjohn@iisc.ac.in}{josyjohn@iisc.ac.in};\href{mailto:velhalb@iisc.ac.in}{velhalb@iisc.ac.in};\href{mailto:vssuresh@iisc.ac.in}{vssuresh@iisc.ac.in}}}

\maketitle
\begin{abstract}
This paper proposes a Conflict-aware Resource-Efficient Decentralized Sequential planner (CREDS) for early wildfire mitigation using multiple heterogeneous Unmanned Aerial Vehicles (UAVs). Multi-UAV wildfire management scenarios are non-stationary, with spatially clustered dynamically spreading fires, potential pop-up fires, and partial observability due to limited UAV numbers and sensing range. The objective of CREDS is to detect and sequentially mitigate all growing fires as Single-UAV Tasks (SUT), minimizing biodiversity loss through rapid UAV intervention and promoting efficient resource utilization by avoiding complex multi-UAV coordination. CREDS employs a three-phased approach, beginning with fire detection using a search algorithm, followed by local trajectory generation using the auction-based Resource-Efficient Decentralized Sequential planner (REDS), incorporating the novel non-stationary cost function, the Deadline-Prioritized Mitigation Cost (DPMC). Finally, a conflict-aware consensus algorithm resolves conflicts to determine a global trajectory for spatiotemporal mitigation. The performance evaluation of the CREDS for partial and full observability conditions with both heterogeneous and homogeneous UAV teams for different fires-to-UAV ratios demonstrates a $100\%$ success rate for ratios up to $4$ and a high success rate for the critical ratio of $5$, outperforming baselines. Heterogeneous UAV teams outperform homogeneous teams in handling heterogeneous deadlines of SUT mitigation. CREDS exhibits scalability and $100\%$ convergence, demonstrating robustness against potential deadlock assignments, enhancing its success rate compared to the baseline approaches.
\end{abstract}

\def\abstractname{Note to Practitioners}
\begin{abstract}
Practical wildfire scenarios often involve unknown clusters of rapidly evolving fires that exceed the available firefighting resources. Wildfire scenarios often involve vast areas and limited sensor capabilities of UAVs, resulting in partial information about the environment. When the number of fires exceeds the number of UAVs, decentralized sequential action becomes necessary. Early wildfire mitigation, focusing on containing fires within the quenching capability of a single UAV, is crucial for minimizing damage and efficient resource utilization by avoiding complex multi-UAV coordination. The approach of single UAV mitigation introduces deadlines for initiating mitigation efforts. The deadlines vary based on factors like fire area, spread rate, and quench rate. The computation of a quenching sequence with efficient prioritization of deadlines ensures mission success and reduction in the total destroyed area. The challenges involved in wildfire scenarios necessitate a three-phased framework: a search stage to locate fires using thermal sensors and cameras, a Resource-Efficient Decentralized Sequential planner to assign local trajectories (mitigation sequence) for each UAV, and a conflict resolution stage to ensure smooth operation by resolving potential conflicts between UAV trajectories. CREDS prioritizes deadlines and achieves successful missions even when fires outnumber UAVs by five times. Additionally, heterogeneous UAV teams with diverse quench and speed capabilities outperform homogeneous teams and traditional methods focused on execution time. This approach is well-suited for scenarios with spatially distributed, dynamic targets with diverse deadlines.
\end{abstract}


\section{Introduction}
\IEEEPARstart{W}{ildfires} pose a significant threat to biodiversity and resource sustainability, particularly when left unchecked in their early stages. Rapid escalation in size and number can occur due to factors like terrain and wind conditions. Emergency response teams face the critical task of decision-making, allocating limited resources to combat fires based on location and severity. Multi-UAV systems are increasingly employed for wildfire management \cite{akhloufi2020unmanned}, \cite{yuan2015survey} to reduce human exposure to hazardous environments. While existing research in the field of multi-UAV wildfire management explores various combinations of search \cite{SARKAR2021}, monitoring \cite{sujit2007fire},\cite{Pham}, and mitigation \cite{Kumar2011}, few studies address the integrated aspects of all three phases \cite{OMS}, \cite{MSCIDC}. The task assignment dimension remains a critical challenge in the realm of multi-UAV wildfire management due to its complex dependencies \cite{korsah2013comprehensive}. This involves assigning UAVs to specific fires based on size and location. Wildfire scenarios are characterized by clustered fire areas that dynamically grow, compounded by the vast geographical area and limited sensing capabilities of UAVs, resulting in limited visibility (partially observable) conditions. Consequently, the multi-UAV multi-task assignment for dynamic tasks remains a complex research area.

Various approaches have been proposed to address the challenge of allocating resources effectively in wildfire scenarios. The problem of scheduling available firefighting resources like fire engines, air tankers, and helicopters to contain multiple fires is solved using the concept of deteriorating jobs in \cite{pappis2010scheduling}. A multi-objective optimization method for fire engine scheduling, by allocating resources to heuristically prioritized fires based on quench capability to travel time, spread rate, and size of fires, is presented in \cite{Tian2016}. A lazy max-sum algorithm for assigning tasks with growing costs to heterogeneous agents is presented in \cite{parker2018lazy}. This approach minimizes the accumulated growth while addressing potential model inaccuracies through a sampling technique. A model-based approach for multi-agent wildfire task assignment to minimize the total completion time is examined in \cite{chen}. This method uses an initial deployment considering the travel time and execution capacity for the task, followed by a dynamic redeployment at intervals based on the optimal execution capacity when travel time is neglected. A tree-based genetic programming hyperheuristic algorithm has been studied in \cite{gao2022mpda} to develop collaborative decision-making strategies to guide multiple robots for bushfire elimination.

While the existing research explores the assignment and scheduling of traditional firefighting resources, a critical gap exists regarding early intervention with limited UAVs under partial observability. The delayed response can escalate fires into Multi-UAV Tasks (MUTs), requiring simultaneous actions and complex coordination strategies. Conversely, managing fires as Single UAV Tasks (SUTs) impose temporal constraints to enable early-stage mitigation while minimizing biodiversity loss and maximizing mission success rates. This strategic approach significantly enhances wildfire management efficiency by reducing resource, computational, and communication demands through streamlined UAV coordination for resource-efficient mitigation. The mitigation of spatially distributed dynamic fires under temporal constraints transforms the problem into a spatiotemporal assignment with dynamic costs. The temporal constraints or deadlines for these growing tasks become heterogeneous due to factors like fire area, fire spread rate, and quench rate of UAVs. Efficient mitigation requires prioritizing these deadlines and utilizing a heterogeneous UAV team with diverse capabilities. The real-world challenges of non-stationary wildfire environments, partial observability, and limited resources often result in uncontrollable fires. This emphasizes the need for decentralized solutions for local planning \cite{liu2023path}. Sequential multi-task assignment strategies \cite{sujit2007sequential} prove beneficial in scenarios with limited agents and non-uniform task distributions in the mission area. Thus, a deadline-prioritized, decentralized, sequential spatiotemporal task assignment for growing fires, utilizing a heterogeneous team of UAVs, is necessary for early-stage wildfire management.

This paper presents the Conflict-aware Consensus-based Resource-Efficient Decentralized Sequential planner (CREDS) for early wildfire management with a heterogeneous UAV team. The CREDS uses a deadline-imposed sequential SUT mitigation to achieve early intervention while reducing the loss of biodiversity. The non-stationary wildfire scenarios with growing fires, new pop-up fires, a limited number of UAVs with partial observability, and heterogeneous deadlines for fires make the multi-UAV wildfire management problem even more challenging. The CREDS tackles the challenge by reformulating the problem as a decentralized, sequential spatiotemporal task assignment problem for growing tasks– an NP-complete problem. CREDS employs a three-phased decentralized framework with path replanning that operates effectively under partial observability, ensuring timely intervention despite incomplete situational awareness. The first phase uses the Oxyrrhis Marina-inspired Search (OMS) algorithm \cite{OMS} to aid UAVs in detecting initially unknown fire locations. The number of fires detected varies with time depending on the search phase, and CREDS replans the path when a new fire is detected. The second phase utilizes a decentralized auction-based task assignment, Resource-Efficient Decentralized Sequential planner (REDS) for planning the local trajectory. The CREDS proposes a Deadline Prioritized Mitigation Cost (DPMC) to efficiently prioritize the tasks based on their heterogeneous deadlines to increase the total number of successful SUT mitigations. Finally, the third phase employs the conflict-aware consensus algorithm to resolve conflicts and generate conflict-free global trajectories for the UAVs. The lack of relevant benchmarks within the existing literature necessitate the creation of a baseline focused on minimizing total execution time to serve as a reference for performance evaluation. The extensive Monte-Carlo simulations validate the effectiveness of CREDS in achieving significantly higher success rates in critical scenarios where fires outnumber UAVs by four to five times, outperforming the baseline cases. The advantage of CREDS is further amplified by heterogeneous UAV teams whose diverse capabilities enable them to tackle fires with heterogeneous deadlines more efficiently than homogeneous teams. Additionally, CREDS consistently achieves $100\%$ convergence with lesser iterations, demonstrating robustness against potential deadlock scenarios, ultimately enhancing its success rate compared to baseline approaches. The resource-efficient task assignment, scalability, and rapid convergence make CREDS a real-time computable solution ideal for timely intervention in real-world wildfire management scenarios. The main contributions of CREDS can be summarized as follows
\begin{enumerate}
    \item An early wildfire management system is proposed to mitigate fires as SUTs before they grow into MUTs, minimizing damage, especially when fires outnumber available UAVs.
    \item A decentralized sequential spatiotemporal task assignment for growing tasks featuring non-stationary costs is used for the computation of mitigation sequence. This facilitates efficient planning and resource utilization even under partial observability. 
    \item CREDS efficiently prioritizes heterogeneous deadlines by utilizing the heterogeneous UAV teams with novel DPMC. Monte-Carlo simulation and ablation studies confirm the ability of CREDS to achieve high success rates, even when fires exceed UAVs by five times.
    \item The CREDS exhibits high success rates, scalability, and faster convergence in resource-limited scenarios, making it suitable for real-world applications with limited resources and growing costs.
\end{enumerate}

The rest of this paper is organized as follows. Section II provides a review of existing multi-robot multi-task assignment approaches. Section III presents the formulation of the wildfire suppression problem as a spatiotemporal task assignment problem. Section IV introduces the CREDS for wildfire management. Numerical simulation results are given in Section V to verify the performance of the CREDS for wildfire suppression, and Section VI concludes the paper.

\section{Related Works}
The wildfire mitigation using a team of UAVs has been framed as a multi-UAV decentralized spatiotemporal task assignment problem in this paper. This section briefly summarizes the existing literature on multi-agent multi-task assignment, covering various problem contexts.

\subsection{Multi-task Assignment Problems}
Multi-Robot multi-Task Assignment (MRTA) has found applications in various domains, including search and rescue \cite{Jin2006searchandrescue}, \cite{zhao2016searchandrescue}, territory protection \cite{Shridhar2022}, job-shop scheduling \cite{Ahn2023jobshop}, vehicle routing problems \cite{Dorling2017routing}, pickup and delivery problems \cite{bai2020routing}, \cite{sajid2022routing}, \cite{Nishida2023delivery}, \cite{Pei2023delivery}, etc. The domain under consideration shapes the characteristics and complexity of the MRTA problem \cite{khamis2015multi}. The Hungarian method \cite{kuhn1955hungarian} is the first approach to efficiently compute optimal solutions for linear sum assignment problems within a finite time among centralized MRTA algorithms. Centralized MRTA systems have drawbacks like scalability, computational complexity, and lack of robustness, making them suitable for scenarios with a small number of robots and tasks with readily available global state information \cite{skaltsis2023review}.

The coordination of a fleet of autonomous vehicles for multi-assignment problems using the Consensus-Based Bundle Algorithm (CBBA) is detailed in \cite{choi2009CBBA}. The CBBA approach adopts a decentralized auction-based strategy for task selection and integrates a consensus routine to resolve conflicts, guaranteeing convergence to conflict-free assignments. An MRTA strategy focused on minimizing mission completion time is detailed in \cite{patel2020decentralized}, employing a decentralized genetic algorithm approach and emphasizing parallelization across agents for efficient computational resource utilization. Building upon the CBBA, \cite{hunt2014CBGA} presents a consensus-based grouping algorithm for coordinating UAVs in MUT assignments considering equipment requirements and task dependencies. In \cite{wang2022consensus}, a consensus-based timetable algorithm addressed the decentralized multi-agent task allocation problem, where multiple agents are needed to perform a task simultaneously. 

\subsection{Multi-task Assignment Problems with Time Constraints} 
MRTA problems involving temporal, spatial, and sequential constraints have diverse applications, ranging from warehouse automation \cite{velhal2023dynamic} to surveillance \cite{Shridhar2022dream}. Challenges include adapting to dynamically discovered tasks, meeting strict arrival requirements, and addressing specific task sequences or concurrent execution needs. An auction-based distributed algorithm for the multi-robot task assignment problem, considering finite task duration, task deadlines, and limited robot battery life, is studied in \cite{Luo2015deadline}. The challenge of maximizing the number of task allocations in a distributed multi-robot system operating under strict time constraints and fuel limits is addressed with Performance Impact-MaxAss (PI-MaxAss) \cite{turner2017distributed}. PI-MaxAss follows a two-phased task assignment strategy where a solution generated from an existing PI algorithm \cite{whitbrook2015novel} is iteratively improved to maximize task assignments without repeating the entire allocation procedure. A distributed task allocation problem for maximizing the total number of successfully executed tasks in multi-robot systems considering task deadlines and fuel limits is studied in \cite{wang2023efficient}. The algorithm minimizes traveling time and prioritizes tasks with earlier deadlines.

Current MRTA literature lacks solutions for managing real-world wildfire scenarios. These scenarios involve the sequential assignment of spatially distributed tasks that exhibit temporal growth (progressively increasing quench times) under partial observability. Additionally, deadlines must be considered for these tasks, and coordinating a heterogeneous team of agents poses a significant challenge. This paper presents a novel CREDS specifically designed for real-world wildfire management to address these shortcomings.

\section{UAV-based Wildfire Management}

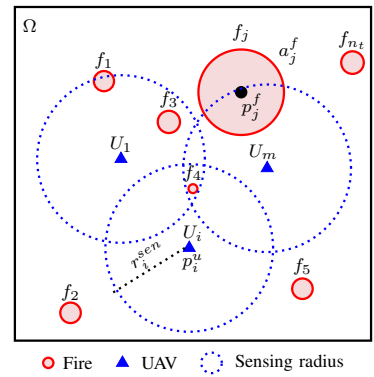
\begin{figure}[!ht]
\centering
\resizebox{0.55\columnwidth}{!}{\begin{tikzpicture}[x=3in,y=3in, scale=0.75]

\draw[line width=0.35mm, color=black] (0.01,0.07) rectangle (0.97,0.97);
\node[inner sep=0.5ex] at (0.05,0.92) {\footnotesize $\Omega$};
\draw[line width=0.35mm, color=red, fill={rgb, 255:red, 245; green, 219; blue, 219 }] (0.25,0.77) circle (5.9pt);
\node[inner sep=0.5ex] at (0.25,0.825) {\footnotesize $f_1$};
\draw[line width=0.35mm, color=red, fill={rgb, 255:red, 245; green, 219; blue, 219 }] (0.16,0.145) circle (6pt);
\node[inner sep=0.5ex] at (0.16,0.2) {\footnotesize $f_2$};
\draw[line width=0.35mm, color=red, fill={rgb, 255:red, 245; green, 219; blue, 219 }] (0.425,0.66) circle (6.5pt);
\node[inner sep=0.5ex] at (0.425,0.72) {\footnotesize $f_3$};
\draw[line width=0.35mm, color=red, fill={rgb, 255:red, 245; green, 219; blue, 219 }] (0.49,0.48) circle (2.6pt);
\node[inner sep=0.5ex] at (0.49,0.522) {\footnotesize $f_4$};
\draw[line width=0.35mm, color=red, fill={rgb, 255:red, 245; green, 219; blue, 219 }] (0.785,0.21) circle (6pt);
\node[inner sep=0.5ex] at (0.785,0.27) {\footnotesize $f_5$};
\draw[line width=0.35mm, color=red, fill={rgb, 255:red, 245; green, 219; blue, 219 }] (0.92,0.82) circle (6.6pt);
\node[inner sep=0.5ex] at (0.92,0.88) {\footnotesize $f_{n_t}$};
\draw[line width=0.35mm, color=red, fill={rgb, 255:red, 245; green, 219; blue, 219 }] (0.62,0.74) circle (25pt);
\node[inner sep=0.5ex] at (0.62,0.9) {\footnotesize $f_{j}$};
\node[mark size=2.5pt,color=black] at (0.62,0.74) {\pgfuseplotmark{*}};
\node[inner sep=0.5ex] at (0.65,0.70) {\footnotesize $p^f_{j}$};
\node[inner sep=0.5ex] at (0.75,0.85) {\footnotesize $a^f_{j}$};

\draw[line width=0.35mm, color=blue, dotted] (0.296,0.56) circle (49pt);
\node[mark size=3pt,color=blue] at (0.296,0.56) {\pgfuseplotmark{triangle*}};
\node[inner sep=0.5ex] at (0.296,0.6) {\footnotesize $U_1$};

\draw[line width=0.35mm, color=blue, dotted] (0.48,0.32) circle (49pt);
\node[mark size=3pt,color=blue] at (0.48,0.32) {\pgfuseplotmark{triangle*}};
\node[inner sep=0.5ex] at (0.49,0.36) {\footnotesize $U_i$};

\node[inner sep=0.5ex] at (0.49,0.28) {\footnotesize $p^u_i$};

\draw[line width=0.35mm,dotted] (0.468,0.32) -- (0.275,0.2);
\node[inner sep=0.5ex, rotate=30] at (0.37,0.295) {\footnotesize $r^{sen}_i$};

\draw[line width=0.35mm, color=blue, dotted] (0.69,0.535) circle (49pt);
\node[mark size=3pt,color=blue] at (0.69,0.535) {\pgfuseplotmark{triangle*}};
\node[inner sep=0.5ex] at (0.68,0.58) {\footnotesize $U_m$};

\draw[line width=0.35mm, color=red, fill={rgb, 255:red, 245; green, 219; blue, 219 }] (0.1,0.01) circle (3pt);
\node[inner sep=0.5ex] at (0.18,0.01) {\footnotesize Fire};
\node[mark size=3pt,color=blue] at (0.3,0.01) {\pgfuseplotmark{triangle*}};
\node[inner sep=0.5ex] at (0.4,0.01) {\footnotesize UAV};
\draw[line width=0.35mm, color=blue, dotted] (0.54,0.01) circle (6pt);
\node[inner sep=0.5ex] at (0.75,0.01) {\footnotesize Sensing radius};
\end{tikzpicture}}
\caption{A wildfire management scenario where multiple UAVs perform search and mitigation of wildfire.}
\label{fig_forestfire_TA}
\end{figure}

A wildfire management scenario featuring multiple UAVs deployed in the mission area, $\Omega\subset \mathbb{R}^2$, to sequentially mitigate randomly located multiple unknown fire areas is shown in Fig. \ref{fig_forestfire_TA}. A team of $m$ UAVs, $\{U_1,..., U_m\}$ is deployed in the mission area to detect and mitigate the set of $n_t$ fires, $\{f_1,..., f_{n_t}\}$. The UAVs need to use the search capabilities to detect the fires and determine individual sequences for efficiently quenching fires, with an overall goal of minimizing biodiversity loss. The active forest fire areas exhibit critical quench rates, beyond which additional quench efforts prove ineffective in reducing the quench time \cite{MSCIDC}. This study assumes UAVs operate with a quench rate close to the critical quench rate, limiting the potential for substantial performance gains with additional UAVs if the fire remains a SUT. The assumptions made in this study include constant fire spread rate, uniform fuel availability, consistent terrain, and wind conditions across the mission area. Additionally, it assumes that UAVs have the sufficient quenching capacity to address multiple fires sequentially, with minimal time required for refueling.

This work focuses on solving the SUT assignment problem, where a single UAV sequentially quenches all detected fires before they escalate into MUT while minimizing forest area destruction. This strategic approach potentially reduces biodiversity loss, eliminates delays associated with waiting for coordinated UAV efforts, and enhances resource utilization by sequentially quenching more fires.

\subsection{Mathematical Formulation}
Let $m$ represent the total number of UAVs, and $\mathcal{U} \triangleq \{1,..., m\}$ denote the index set of UAVs. The UAVs utilized in the mission exhibit heterogeneity, possessing distinct speeds and quenching capabilities. The position coordinate of the $i$-th UAV is denoted by $p^u_{i}$. The speed in \si{m/s} and the area quench rate in \si{m^2/s} for the $i\text{th}$ UAV are denoted as $v_i$ and $\phi^q_i$ respectively.
Each UAV denoted as $U_i$, is confined to sensing only the fire locations within its designated sensing radius, $r^{sen}_i$. The detected fire locations are stored in a dedicated list denoted as $\mathcal{D}_i$. The UAVs navigate within a partially observable environment where the initial details regarding the number, size, and locations of fires are unknown.

In a typical wildfire scenario, the tasks or fires are dynamic in terms of size and number as the fire particles can be carried away due to the wind and terrain conditions, resulting in an increased number of clustered fire areas. Let $n_t$ represents the number of fires in the mission area at time $t$ in seconds, with the assumption that the number of fires is higher than the number of deployed UAVs for the mission ($n_t >> m$). The index set of fires is denoted as $\mathcal{F}\triangleq \{1,..., {n_t}\}$ and the center of the $j\text{th}$ fire spot is located at the position coordinate $p^f_{j}$. Each fire, $f_j$ in the cluster of fires, can be modeled as a point fire model with a circular fire profile \cite{chen}. The initial areas of the fires are assumed to be heterogeneous with different initial areas and $a^f_{j0}$ be the initial area of $f_j$ in \si{m^2}. The fire area grows with time, depending on the radial spread rate, $\phi^s_j$ in \si{m/s}. Let $a^f_j(t)$ and $P^f_j(t)$ be the area in \si{m^2} and perimeter in \si{m} of the $f_j$ at $t$ \si{s} respectively. The perimeter can be expressed as a function of fire area as $P^f_j(t)=2\sqrt{\pi}\sqrt{a^f_j(t)}$. When $U_i$ detects and acts on $f_j$, the rate of change of the area of $f_j$ can be expressed as a first-order system equation as:

\begin{gather}
    \dot{a}^f_j(t)=\phi^s_jP^f_j(a^f_j)-\phi^q_i.
    \label{eq_firearea}
\end{gather}

The fire area decreases when $\dot{a}^f_j(t)<0$, specifically when $\phi^q_i>\phi^s_jP^f_j(a^f_j)$. The critical area, denoted as $a^{c}_{ij}$, is the threshold fire area below which the fire remains as a SUT for $U_i$. The critical area, $a^{c}_{ij}$ can be computed as:
\begin{gather}
    a^{c}_{ij}=\Big(\frac{\phi^q_i}{2\sqrt{\pi} \phi^s_j}\Big)^2 ~~ i\in \mathcal{U}, j\in \mathcal{D}_i. 
\end{gather}
If $a^f_j>a^{c}_{ij}, ~\forall i$, then the fire area exceeds the critical area, rendering the task infeasible for a single robot in the team. The critical time at which the fire area grows to $a^{c}_{ij}$ represents the deadline time for successful mitigation of $f_j$ by $U_i$ as a SUT. The deadline time of $f_j$ for $U_i$ is denoted as $\tau^{d}_{ij}$ and can be computed as:
\begin{gather}
    \tau^d_{ij}=\frac{\sqrt{a^{c}_{ij}}-\sqrt{a^f_{j0}}}{\phi^s_j\sqrt{\pi}}~~ i\in \mathcal{U}, j\in \mathcal{D}_i. 
\end{gather}
The critical area and deadline time are heterogeneous for UAVs depending on the initial fire area, fire spread rate, and quench rate of the UAV acting on the fire. Each UAV has distinct observations, and given the detected fire list for individual UAVs and the deadline time for their detected fires, this work aims to assign a sequence of fires to each UAV strategically. The mission is considered successful when all fire areas are effectively mitigated as SUTs, meaning no fires become infeasible due to missed deadlines.

The path of $U_i$, denoted as $\mu_i$, is the ordered sequence of fires to be mitigated. The paths play an important role in the success of the mission, and the path computation is challenging for complex dynamic environments like wildfire scenarios. The path computed should ensure that all UAVs commence mitigation of fires in their path before the respective deadline times while minimizing the loss of biodiversity. The primary objective of computing paths of UAVs is to achieve successful mission through resource-efficient SUT mitigation of all fires in the mission area. Mathematically, this translates to a constraint ensuring that all fires in the paths of all UAVs are addressed before their individual deadlines. This can be expressed as:
\begin{gather}
    \tau^s_{ij}-\tau^d_{ij}<0~~ \forall j\in \mu_{i}, \forall i\in \mathcal{U}, 
    \label{eq_SUTcondition}
\end{gather}
where, $\tau^s_{ij}$ is the start time of mitigation of $f_j$ by $U_i$. The start time includes the quench times for the previous tasks in the path and the travel time to reach the current task from the initial position along the path. Let the path, $\mu_i=\{\mu_{i1},\mu_{i2},...,\mu_{il},...\}$ be executed by $U_i$. The element, $\mu_{il}$, denotes the task executed by $U_i$ at the $l\text{th}$ point of the path. The $U_i$ should search, detect, travel, and start quenching the fire before that fire becomes an infeasible quenching task for $U_i$. The $U_i$ should arrive at the fire location, $\mu_{il}$ before $\tau^d_{i\mu_{il}}$ to limit the fire area below $a^{c}_{i\mu_{il}}$ to mitigate the fire as a SUT, i.e $\tau^s_{i\mu_{il}}<\tau^d_{i\mu_{il}}$. The secondary objective of path computation is to minimize the total biodiversity loss, achievable through minimizing the start time of mitigating fires in the UAV's path. Computing $\mu_{i}$ with a focus on minimizing the start time, while satisfying \eqref{eq_SUTcondition}, ensures that all fires are quenched as SUTs at an early stage, reducing biodiversity loss and maximizing resource efficiency. 

In real-time evolving wildfire environments with dynamic task execution costs, a real-time computable, scalable, and resource-efficient solution is required. Each UAV individually selects the fires to mitigate in a decentralized manner based on the start time of mitigation of fires using a resource-efficient decentralized sequential planner, and a conflict-aware consensus algorithm is employed for the conflict resolution.

\section{Development of CREDS}

\begin{figure}[t]
\centerline{\includegraphics[width=75mm]{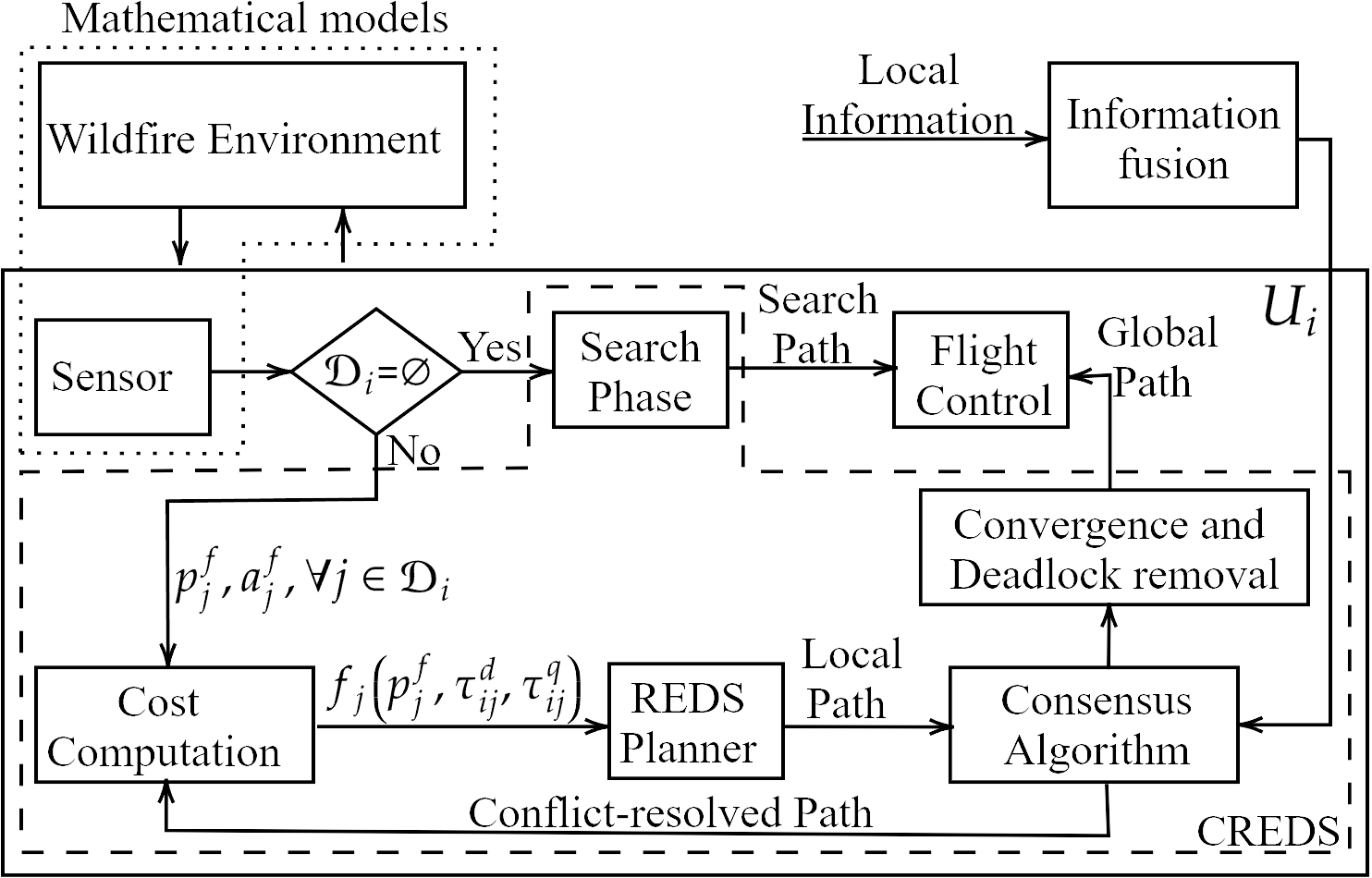}}
	\caption{Schematic diagram of the CREDS for a UAV.}
	\label{fig_TA_block}
\end{figure}

The Conflict-aware Resource-Efficient Decentralized Sequential planner is designed to compute sequential assignments for all UAVs involved in wildfire management missions, ensuring the timely quenching of all fire areas before their respective deadline times. The planner prioritizes efficient resource utilization and aims to minimize the overall area destroyed by fires. The schematic diagram depicting the components of the CREDS for a UAV, $U_i$, is shown in Fig. \ref{fig_TA_block}. Wildfire management comprises three phases: search, path generation, and conflict resolution. All UAVs are equipped with temperature sensors that measure the temperature and temperature gradient required for the search phase. The UAVs also have an infrared camera that senses the fire profile parameters when the fire location is within the sensing radius. Due to the limited sensing range, each UAV will have only local observations. The UAVs individually search for fires using OMS \cite{OMS} to detect the fires.

The CREDS adopts an auction-based methodology that employs cost functions to determine the paths of the UAVs. Each UAV employs the Resource Efficient Decentralized Sequential planner to generate the path of mitigation, utilizing information on detected fires and mission objectives. The formulated cost function should account for both the deadline time and start time of mitigation, reflecting the total destroyed area to achieve mission objectives. The REDS works in a decentralized manner to minimize the individual costs of UAVs, and the same task could be added to the paths of multiple UAVs. The conflict resolution phase uses a conflict-aware consensus algorithm to compute a conflict-free path by assigning fires to UAVs, ultimately minimizing the overall mission cost. The assigned UAV should continuously spray water on the firefront for the scheduled quench time to quench the fire completely. Each UAV can quench only one fire at a time, and after complete mitigation, the UAV moves to the next task in the path. If the path is an empty set, the UAV needs to search and detect fires in unexplored areas. The task execution is activated when the UAVs start the traversal from their previous task to the current task. When a new fire is detected, UAVs with activated tasks cannot alter their current task, but their upcoming path is recomputed, considering the existing tasks in the detected fire list.

\subsection{Search Phase}
The search phase of fires is required as the UAVs have only partial information about the fires present in the mission area. The UAVs can detect only the fire areas located within their sensing radius, and the detected fire locations of UAV, $U_i$, are denoted by set $\mathcal{D}_i=\{j, ~\text{if}~ d_{ij}<r^{sen}_i,~\forall j \in \mathcal{F}\}$, where $d_{ij}$ is the distance from the $U_i$ to $f_j$. The UAVs individually search for fire areas using OMS if the corresponding detected fire list, $\mathcal{D}_i=\varnothing$. The OMS is a multilevel search inspired by the foraging pattern of Oxyhrris Marina. The UAVs use temperature and temperature gradient measurements to switch between the explorative and exploitative search levels. The UAVs use levy search to explore the area when the temperature is lower than the temperature threshold value. The UAVs employ the exploitative brownian search if the temperature is higher than the temperature threshold. Further, if the temperature gradient is positive, a directionally driven brownian search with a smaller search space is employed to reach the accurate fire location \cite{OMS}. The fire is detected when the fire location is within the sensor radius of the infrared camera, and the fire is added to the detected fire list of detected UAV. Thus, for a fire $j \in \mathcal{D}_i$, the detected UAV will have the position coordinates $p^f_j$, fire area $a^f_j$, and the spread rate $\phi^s_j$ of the fire.

\subsection{Cost Function Formulation}
The formulation of cost functions is essential in the context of the CREDS, which computes the fires being quenched by each UAV and determines the order in which each UAV mitigates the fire spot to minimize the overall cost to the team of UAVs. The multi-task assignment problems dealing with static tasks typically employ cost functions such as the total distance covered or the total mission time. Unlike static tasks, wildfires exhibit dynamic characteristics, necessitating a cost function that accommodates the growing size of the fire areas and their mitigation requirements. Despite the clustering of fires, travel time remains crucial, particularly as the size of the fire area expands when the UAV reaches the fire location. The feasibility of a task is contingent on whether the fire area exceeds the critical area or if the start time of mitigation exceeds the deadline time.

Let $\tau^s_{ij}$ represent the start time of mitigation of $f_j$ by $U_i$. The start time of mitigation, $\tau^s_{ij}$, is computed as:
\begin{gather}
		\tau^s_{ij}=
  	\begin{cases}
		\frac{d_{ij}}{v_i},& l=1;\\
		\tau^c_{ij'}+\frac{d_{j'j}}{v_i},& l>1.
	\end{cases}
\end{gather}
where $l$ denotes the position in the path, $d_{ij}$ is the distance from the initial position of $U_i$ to $f_j$ if $j$ is the first task assigned to $U_i$, $j'$ is the previous task assigned to $U_i$, $\tau^c_{ij'}$ is the time of completion of the previous task, and $d_{j'j}$ is the distance from the previous task to the current task. The start time of mitigation of tasks in the path depends on the completion time of preceding tasks in the path. The completion time of a task is defined as the time at which a fire is completely quenched. The completion time of a task in the path is the summation of individual task execution times up to the task in the path. The execution time of each task includes the travel time and quench time required for the corresponding fire area.

The quench time is the time required to quench the fire after the start of mitigation. The quench time of the fire depends on the size of the fire area, the spread rate, and the quench rate, considering a specific number of UAVs acting on the fire. The size of the fire area to be quenched alternatively depends on the start time of mitigation. Let $\tau^q_{ij}$ represents the quench time for SUT mitigation of $f_j$ by $U_i$. The value of $\tau^q_{ij}$ for feasible tasks can be computed from the solution for \eqref{eq_firearea} assuming fire area at start time, $a^f_j(\tau^s_{ij})$ as the initial condition, and the fire area after quench time as $a^f_j(\tau^s_{ij}+Q^t_{ij})=0$ \cite{chen}. The $\tau^q_{ij}$ can be computed as,
\begin{gather}
\hspace{-0.8em}
		\tau^q_{ij}= 
            \begin{cases}
              \frac{2\phi^q_i}{(K\phi^s_j)^2} \text{ln}\Big(\frac{\phi^q_i}{\phi^q_i-K\phi^s_ja^f_j(\tau^s_{ij})}\Big)-\frac{2\sqrt{a^f_j(\tau^s_{ij})}}{K\phi^s_j} ,\tau^s_{ij}<\tau^d_{ij};\\
              \infty ,\tau^s_{ij}\ge\tau^d_{ij}.
            \end{cases}
            \hspace{-1.85em}
            \label{eq_Qtij}
\end{gather}
where $a^f_j(\tau^s_{ij})$ is the fire area of $f_j$ when $U_i$ starts the mitigation and $K=2\sqrt{\pi}$. The quench time for SUT assignment is a finite value as in \eqref{eq_Qtij} for all feasible SUTs and $\infty$ for all infeasible tasks. The quench time for a particular task depends on the time of the start of the task, which is influenced by the execution times for the previous tasks. Let $\tau^e_{ij}$ be the time required to execute $f_j$ by $U_i$ and $\tau^e_{ij}$ is calculated as,
\begin{gather}
		\tau^e_{ij}=
            \begin{cases}
		\frac{d_{ij}}{v_i}+\tau^q_{ij}~,& l=1 \text{~and~} \tau^s_{ij}< \tau^d_{ij};\\
		\frac{d_{j'j}}{v_i}+\tau^q_{ij}~,& l>1 \text{~and~} \tau^s_{ij}< \tau^d_{ij};\\
            \infty ,& \tau^s_{ij}\ge \tau^d_{ij}.
            \end{cases}
\end{gather}

The execution time is a finite value for all SUTs and $\infty$ for all MUTs. The completion time denoted as $\tau^c_{ij}$ marks the time when $f_j$ is completely quenched by $U_i$. The completion time of task $\mu_{il}$, at the $l\text{th}$ point of the path, is the summation of execution times of all tasks up to the $l\text{th}$ point of the path. 
\begin{gather}
    \tau^c_{i\mu_{il}}=\sum_{j \in \{\mu_{i1},...,\mu_{il}\}} \tau^e_{ij}.
\end{gather}
The start time, completion time, and execution time of tasks are all interdependent, and each task's feasibility is intricately linked to the start times of preceding tasks in the path.

\subsubsection{Deadline-Prioritized Mitigation Cost}
The CREDS introduces a novel cost function, Deadline-Prioritized Mitigation Cost, to achieve the objective of maximizing the SUT assignments while minimizing the total destroyed area. The DPMC has two parts: temporal deadline cost and start time cost. The temporal deadline cost ensures the maximization of SUT assignments, and the start time cost ensures the minimization of the total destroyed area. The temporal deadline cost accounts for the difference in area at the time of start to the critical area. The start time cost is a dynamic duration-dependent travel cost that includes the cost of quenching previous tasks in the path and the travel cost to reach the current task from the initial position along the path. A UAV can perform multiple tasks sequentially, and the total score for $U_i$, denoted as $S_i(\mu_i)$, to perform tasks in the feasible path $\mu_i$ using DPMC can be computed as, 
\begin{gather}
 S_i(\mu_i)=\underbrace{\sum_{j \in \mu_i}\Big(\sqrt{a^c_{ij}}-\sqrt{a^f_j(\tau^s_{ij})}\Big)}_{\text{Temporal deadline cost}}.\underbrace{\sum_{j \in \mu_i} \tau^s_{ij}}_{\text{Start time cost}}.
 \label{eq_DPMC_obj}
\end{gather}
The $S_i(\mu_i)=\infty$ for infeasible paths where the start time is greater than the deadline time. The mitigation of fire should start before the deadline, and sometimes fires may not be mitigated before the deadline as UAVs will be busy mitigating others. The temporal deadline cost is minimal when the fires in a feasible path are mitigated closer to their deadline times. The minimization of temporal deadline cost leads to later mitigation of fires with later deadlines. This cost function creates time slots for fires with earlier deadlines and maximizes the number of fires that can be executed as SUTs. 

The key idea for minimizing the destroyed area is to minimize the start time of mitigation to reduce the spread of fire. Due to the dynamic quench time associated with fires, if larger fires with earlier deadlines and higher quench times are mitigated first, it might lead to an increase in the fire area of other fires, leading to more destroyed areas. The temporal deadline cost alone is not sufficient for minimizing destroyed areas for evolving tasks with dynamic costs, as it might push the start time of tasks to deadline times. Thus, both the temporal deadline cost and start time cost are required to be minimized to maximize SUT assignments while minimizing the total destroyed area. The DPMC accounts for the dynamic costs of fires as a product of both temporal deadline cost and start time cost along the path to prioritize the deadlines effectively. The CREDS effectively balances resource efficiency with deadlines to minimize fire damage by minimizing both the temporal deadline and the start time cost.

\subsubsection{Baseline}
In the absence of established benchmarks for this specific multi-UAV wildfire management scenario, a baseline approach to minimize the total execution time is established within the same CREDS framework for the comparative evaluation. The total execution time cost is the time taken to completely quench all the fire areas. The total score for $U_i$ performing tasks in the path $\mu_i$ is the sum of execution time for all tasks in the path, $\mu_i$, and can be computed as, 
\begin{gather}
    S_i(\mu_i)=\sum_{j \in \mu_i} \tau^e_{ij}.
    \label{eq_comp_obj}
\end{gather}

The cost function in \eqref{eq_comp_obj} is used as the baseline for comparison with the performance of CREDS with \eqref{eq_DPMC_obj}. A new task is inserted into the location that incurs the minimum score improvement, and the marginal score improvement associated with task $j$ given the current path, $\mu_i$ is
\begin{gather}
	C_{ij}=\min_{\eta\le|\mu_i|}S_{i}^{\mu_i\oplus_\eta{\{j\}}}-S_{i}^{\mu_i}, ~~\forall j \notin \mu_i.
 \label{eq_local_reward}
\end{gather}
where $\mu_i\oplus_\eta{\{j\}}$ denotes task $j$ is added after $\eta\text{th}$ component of path $\mu_i$. The local reward associated with the new task is the marginal score improvement associated with the task.

\subsection{Spatiotemporal Mitigation for Evolving Wildfires}
The spatiotemporal task for $U_i$ mitigating $f_j$ can be defined as $F_j(p^f_j, \tau^d_{ij},\tau^q_{ij})$ where the UAV has to reach the fire location $p^f_j$ to start the mitigation before $\tau^d_{ij}$ and spray the water for $\tau^q_{ij}$ to quench the fire completely. The detected task set of $U_i$ is defined as $T^u_i=\{F_j ~\forall j \in D_i\}$. The information set of $U_i$ can be defined as $I^u_i=\{p^u_i,T^u_i\}$. The path, $\mu_{i}$ of $U_i$, is a function of the start time and scheduled quench time for fires in the path. The start time of all fires in the path of $U_i$ is defined as $\tau^s_{i\ast}=\{\tau^s_{ij},~\forall j \in \mu_{i}\}$. The scheduled quench time of all fires in the path of $U_i$ is defined as $\tau^q_{i\ast}=\{\tau^q_{ij},~\forall j \in \mu_{i}\}$. 
Thus, $\mu_{i}=g_1(\tau^s_{i\ast},\tau^q_{i\ast}),~ \tau^q_{i\ast}=g_2(\mu_{i},\tau^s_{i\ast}),~\tau^s_{i\ast}=g_3(\mu_{i},\tau^q_{i\ast})$, where $g_1$, $g_2$ and $g_3$ are functions. The quench time or scheduled time for fire depends on the fire area at the time of start and the area quench rate of the UAV, which in turn depends on the path of the UAV and the start times of the mitigation of fires in its path. The start times of the fires in the path depend on the time to reach the fire locations and the completion times of previous fires in the path, which alternately is a function of the path and quench time of all fires in the path.

The wildfire management problem can be considered as a superset of the technician routing and scheduling problem with repair times, which belongs to the class of NP-Complete problems \cite{pillac2013TRSP}, \cite{afrati1986TRP}. The processing (repair) time for a task is constant for the technician routing and scheduling problem with repair times, whereas the quench time of fires gradually increases with time. The growing nature of the fire demands dynamic (and increasing) quench time, and this makes it an NP-complete problem.

The spatiotemporal mitigation for wildfire tasks needs the UAVs to reach the fire area and mitigate the fire before it becomes a MUT, indirectly enhancing resource utilization. The UAVs should sequentially mitigate all fire locations when the fire area remains as a SUT for a successful mission. The spatiotemporal mitigation for wildfire management can be formulated as a multi-task assignment problem given by,
\begin{gather}
    \label{eq_perfobj}
    \min_{x_{ij}}~ \sum_{i=1}^{m} \Bigg(\sum_{j=1}^{n_t} C_{ij}(\mu_i)x_{ij}\Bigg),\\
    \label{eq_SUT_alloc}
    \text{s.t}~ \sum_{i}^{m}x_{ij} \le 1~~\forall j \in \mathcal{F},\\
    \label{eq_temporal_constraint}
    \tau^s_{ij}-\tau^d_{ij}<0 ~~\forall j \in \mu_i, \forall i\in \mathcal{U},\\
    \label{eq_bin_var}
    x_{ij} \in \{0,1\} ~~\forall (i,j) \in \mathcal{U} \times \mathcal{F},\\
    \label{eq_bin_PO}
    x_{ij}=0 ~~\forall j\notin \mathcal{D}_i.
\end{gather}
The sequential spatiotemporal task assignment computes the paths of UAVs, minimizing the total cost of task execution in \eqref{eq_perfobj} while guaranteeing a successful mission. The task assignment problem is formulated as an integer programming problem to compute the optimal path $\mu_i,~\forall i\in \mathcal{U}$. The UAVs can execute multiple tasks sequentially, and the path $\mu_i$ contains the ordered sequence of tasks assigned to $U_i$. The SUT mitigation condition where each fire area is assigned to a single UAV is given in \eqref{eq_SUT_alloc} and the temporal constraints to quench the fire with a single UAV is given in \eqref{eq_temporal_constraint}. The binary decision variable, $x_{ij}=1 ~\forall j\in \mu_i$ and $0$ otherwise, is given in \eqref{eq_bin_var}. The partial observability condition, where the fires get assigned to UAVs only if it is observable to the UAVs, is added as \eqref{eq_bin_PO}.

The cost function, $C_{ij}$, is a path-dependent reward for the execution of the task, $F_j$ by $U_i$. In this work, $C_{ij}$ is computed using both DPMC and baseline cost to analyze the effect of both cost functions. The sequential assignment of spatially located growing tasks with temporal constraints and dynamic processing times presents significant computational challenges in finding optimal solutions. Traditional optimization methods applied to \eqref{eq_perfobj} for wildfire task assignment in partially observable environments face computational complexity due to the NP-complete nature of the problem. The spatiotemporal assignment of tasks for UAVs in wildfire management involves complex dependencies \cite{korsah2013comprehensive}, where the mission's success relies heavily on the coordinated actions and schedules of the UAV team. Given the tight coupling between individual tasks and overall path planning, sophisticated coordination strategies are essential to optimize resource allocation and achieve effective fire mitigation. The proposed CREDS addresses this complexity by employing a decentralized auction-based approach.

\subsection{Resource Efficient Decentralized Sequential Planner}
The Resource Efficient Decentralized Sequential planner uses an auction-based algorithm for the spatiotemporal SUT assignments to find a conflict-free task assignment for detected fires. In a conflict-free SUT assignment, each task is assigned to at most one UAV. The REDS determines the local path for each UAV in a decentralized manner, using DPMC. The UAVs adopt a greedy approach to augment their path lists from the set of detected fire areas, $D_i$, to minimize local or individual rewards. The local reward for $U_i$ for executing all assigned tasks is $\sum_{j}^{n_t} C_{ij}(\mu_i)x_{ij}$.

The algorithm of the REDS is summarized in Algorithm \ref{alg:one}. In REDS, each UAV, $U_i$, stores and updates $4$ vectors, bundle ($b_i$), path ($\mu_i$), winning bids ($y_i$), and winners list ($z_i$) . The bundle is a list of tasks in the order of addition, and the path is the sequential order of execution of tasks. The winning bids contain the lowest bid for all the tasks, and the winners are the UAVs with minimum bids. The bundle and path are initialized as null vectors, the winner list is initialized to zero vector, and winning bids are initiated as $\infty$ for all tasks. For iterations, $\lambda>1$, bundle, path, winner bids, and winners are initiated as in the previous $(\lambda-1)\text{th}$ iteration. The UAVs compute local rewards for all tasks in their detected task list using \eqref{eq_local_reward} and compute their valid task lists, $h$. The indicator function, $\mathbb{I}(·)$, compares the local reward with the winning bid. $\mathbb{I}(·)$ is 1 if the conditional argument is true and $\infty$ otherwise. If the computed local reward is less than the known winning bid for the task, then the task is valid, and $h_{ij}=1$; otherwise, the task is invalid and $h_{ij}=\infty$. The addition of valid tasks in the path minimizes local rewards for UAVs, and the UAV has a higher probability of winning valid tasks as the individual reward gain is less than the existing winner. Hence, the UAVs bid for only valid tasks, and the task with a minimum reward, $J_i$, is added to the end position in the bundle. The task is added to the position in the path, $\eta_{i, J_i}$, which gives the minimum total score for the UAV. The winner bids and winners are updated according to tasks added to the path.

The REDS tackles a minimization problem, where the task with the lowest reward will be greedily added to the bundles and paths of UAVs iteratively. The goal is to achieve a minimum score for tasks added to the path while determining a feasible path. In baseline cost, UAVs prioritize tasks with shorter completion times i.e., near smaller fires from the detected list are initially added to the bundle. However, larger tasks with higher bids tend to be added as the last tasks, potentially resulting in infeasible paths and infinite costs. This tendency may eventually lead to the exclusion of large fires from the paths.

The REDS prioritizes the addition of tasks with closer deadlines in the path. The large fires with earlier deadlines will have smaller bids with DPMC and are greedily added as the first tasks to bundles, while tasks with later deadlines are included in subsequent iterations. This strategic approach ensures the addition of distant larger fires in the path while allowing near smaller fires to be suboptimally placed in the paths of other UAVs, ensuring the generation of feasible paths. The larger fires are not prioritized as the first task in the path unless their deadline is imminent, rendering them infeasible if there are any preceding tasks. The addition of larger fires at the beginning of the path impacts the start time of subsequent tasks, leading to an increase in start time costs and reducing the slots for feasible tasks in the path. Additionally, the prolonged quench time for the large fire may allow smaller tasks to grow into infeasible ones before the start of mitigation efforts. This explains the significance of simultaneously minimizing the product of temporal deadline cost and start time cost within the DPMC.

The balance between minimizing temporal deadline costs and start time costs is crucial for devising effective paths that adhere to deadlines while minimizing associated start time costs. For both baseline cost and DPMC, the newly added tasks in the path may become infeasible if $\tau^s_{ij}\ge \tau^d_{ij}$, making the local reward of the UAVs $\infty$. The auction-based REDS plays a crucial role in adding newly introduced tasks to positions on the path where tasks will be mitigated before the deadline time while generating a minimum finite local reward for UAVs. However, the DPMC demonstrates a higher likelihood of obtaining feasible paths by prioritizing larger fires, thereby enabling the strategic integration of smaller fires into the paths of UAVs.

\begin{algorithm}[t]
\DontPrintSemicolon
	\caption{Algorithm of REDS}\label{alg:one}
        \While {\textup{non-converging}} { 
        \For {$i=1\text{ to }n_{A}$}{
		Initialize with $m$ UAVs with $D_i$\;
        Initialize $b_i(\lambda)=b_i(\lambda-1),\mu_i(\lambda)=\mu_i(\lambda-1),y_i(\lambda)=y_i(\lambda-1),z_i(\lambda)=z_i(\lambda-1)$\;
        \eIf{$D_i\neq \varnothing$}{		
        
        $C_{ij}=\min_{n\le|\mu_i|}S_{i}^{\mu_i\oplus_\eta{\{j\}}}-S_{i}^{\mu_i} \forall j \in D_i\setminus b_i(\lambda)$\;
        $h_{ij}=\mathbb{I}(C_{ij}<y_{ij}(\lambda))$\;
        $J_i=\argmin _j C_{ij}.h_{ij}$\;
        $\eta_{i,J_i}=\argmin _j S_{i}^{\mu_i\oplus_\eta{\{j\}}}$\;
        $b_i(\lambda)=b_i(\lambda)\oplus_\text{end}{J_i}$\;
        $\mu_i(\lambda)=\mu_i(\lambda)\oplus_{\eta_{i,J_i}}{J_i}$\;
        $y_{i,J_i}(\lambda)=C_{i,J_i}$\;
        $z_{i,J_i}(\lambda)=i$}{Oxyhhris Marina Inspired Search}}
        Conflict-Aware Consensus Algorithm (Call Algorithm 2)}
\end{algorithm}
\begin{algorithm}[t]
\DontPrintSemicolon
 	\caption{Conflict-Aware Consensus Algorithm}\label{alg:two}
        \While {\textup{non-converging}}{
        \For{$i=1\text{ to }n_{A}$}{
        Initialize $b_i(\lambda)=b_i(\lambda-1),\mu_i(\lambda)=\mu_i(\lambda-1),y_i(\lambda)=y_i(\lambda-1),z_i(\lambda)=z_i(\lambda-1),T=1$\;
        {REDS Algorithm}\;
        \If{$g_{ik}=1$}{
        \For{$j=1\text{ to }n_{f}$}{
        \If{$z_{kj}(\lambda)=k$}{\If{$z_{ij}(\lambda)=k$}{Update}
        \ElseIf{$z_{ij}(\lambda)=0$}{Update}
        \ElseIf{$y_{kj}(\lambda)<y_{ij}(\lambda)$}{Update}}    
        \ElseIf{$z_{kj}(\lambda)=i$}{\If{$z_{ij}(\lambda)=k$}{Reset}}          
        \ElseIf{$z_{kj}(\lambda)\notin \{i,k\}$}{\If{$z_{ij}(\lambda)=k$}{Reset}}
        \ElseIf{$z_{kj}(\lambda)=0$}{\If{$z_{ij}(\lambda)=k$}{Update}}}}}
        \If{$z_i(\lambda)=z_k(\lambda) ~\forall~i,k$ \textup{for $w_1$ iterations}}{Algorithm converges}
        \Else
        {Continue $w_2$ iterations- deadlock removal phase}
        {Select assignment with minimum infeasible tasks}\;
        }
        {Update:$y_{ij}=y_{kj},z_{ij}=z_{kj}$}\;
        {Reset:$y_{ij}=\infty,z_{ij}=0$}
\end{algorithm}
\subsection{Conflict-Aware Consensus Algorithm}
The local path generated for UAVs using REDS, may not be conflict-free as each UAV greedily adds tasks to its local path. The UAVs share the information based on the communication network, and the UAVs communicate their winner bids, $y_i$, and winners list, $z_i$, with other UAVs synchronously. The consensus rules are employed for conflict resolution among the UAVs during each iteration, where the conflict tasks are assigned to UAVs with a minimum bid (local reward), and winning data gets updated in each UAV. The consensus rules are summarized in Algorithm \ref{alg:two}. After consensus, UAVs may lose some tasks in their local path, and the tasks get unassigned. The tasks added in the bundle after the unassigned task become invalid as the local reward is path-dependent. Therefore, the UAVs must discard all the tasks added after the unassigned task, and a conflict-resolved path is computed. The stored vectors in the UAVs, $b_i$, $\mu_i$, $y_i$, and $z_i$, are updated depending on the discarded tasks. 

The conflict-aware consensus algorithm provides conflict resolution and global path generation, minimizing global rewards in \eqref{eq_perfobj} by assigning conflict tasks to the UAV having minimum individual rewards. The UAV recomputes the local rewards for the observed tasks that are not in the trajectory from its new initial position, and the process is repeated until the algorithm finally converges to a conflict-free global path. The consensus algorithm converges when the conflict-free global path remains unchanged for future iterations. In this work, if the task winners remain the same for $w_1$ iterations, the algorithm is assumed to be converged. 

The non-stationary nature of the problem, together with the heterogeneous deadlines and UAVs, may lead to a deadlock condition where the assignment toggles at each iteration, leading to non-convergence for certain cases. The deadlock condition arises in the baseline cost function due to the higher costs associated with critical large fires. Consequently, these critical large fires might either be assigned to UAVs as the final task added to the path or excluded from the path, making it infeasible and thereby increasing the occurrence of deadlock scenarios. In such cases, a deadlock removal mechanism is employed where the consensus algorithm is run further for $w_2$ iterations, and the solution having minimum infeasible tasks is selected to declare the winners and the winning bids. 

\section{Results and Discussion}
The performance of CREDS is evaluated for homogeneous and heterogeneous teams of UAVs operating in both fully and partially observable cases for dynamic numbers and sizes of fires. In instances where no fire locations are detected within the sensing range, the UAVs employ the OMS method for searching fire areas. In this study, different UAVs are assumed to fly at different altitudes. The fire areas possess a circular profile and exhibit initial area heterogeneity. The performance of CREDS is compared with the baseline cost. The simulations are conducted in the MATLAB R$2022$b environment, utilizing an Intel Core-i$7$, $3.2-$GHz processor, and $16-$GB memory. The simulation results for a typical scenario to explain the working of the CREDS for wildfire management are provided in supplementary material. A detailed Monte Carlo study is presented next to evaluate the performance of the CREDS under different scenarios.

\subsection{Monte-Carlo Analysis}
A Monte Carlo analysis is conducted to assess the average performance of the CREDS under dynamic conditions encountered in wildfire management. A team of five UAVs operating in a search area of size, $1$ \si{km} $\times$ $1$ \si{km} to search and quench varying numbers of fires is considered. The initiation of the search and detection process occurs randomly from different locations within the considered search area, with UAVs detecting fire locations within the $300$ \si{m} sensing radius. Cases involving $15$, $20$, and $25$ fires located in the search area, with initial radii randomly selected between $5$ to $15$ \si{m}, are studied. A Monte-Carlo analysis, comprising $100$ iterations for each case study, assesses the average performance of the CREDS across randomly initialized fire radii and UAV initial locations for fixed fire center coordinates.

The performance indices used to evaluate the Monte-Carlo simulation are the success rate, mean completion time, mean quench time, and mean Fire Expansion Ratio (FER). The percentage of successful simulations among the total number of simulations is the success rate. The mean completion time is the average value of time required to detect and completely quench all the fires in the search area. The mean quench time is the average of the total quench time required to quench the fires in the search area. The FER is the ratio of the change in the area of fire after the start of the mission to the initial area of fire at the start of the mission. The mean FER accounts for the additional destroyed area after the start of the mission \cite{MSCIDC}. 

\begin{figure}[t]	
    \centerline{\includegraphics[width=8.8cm]{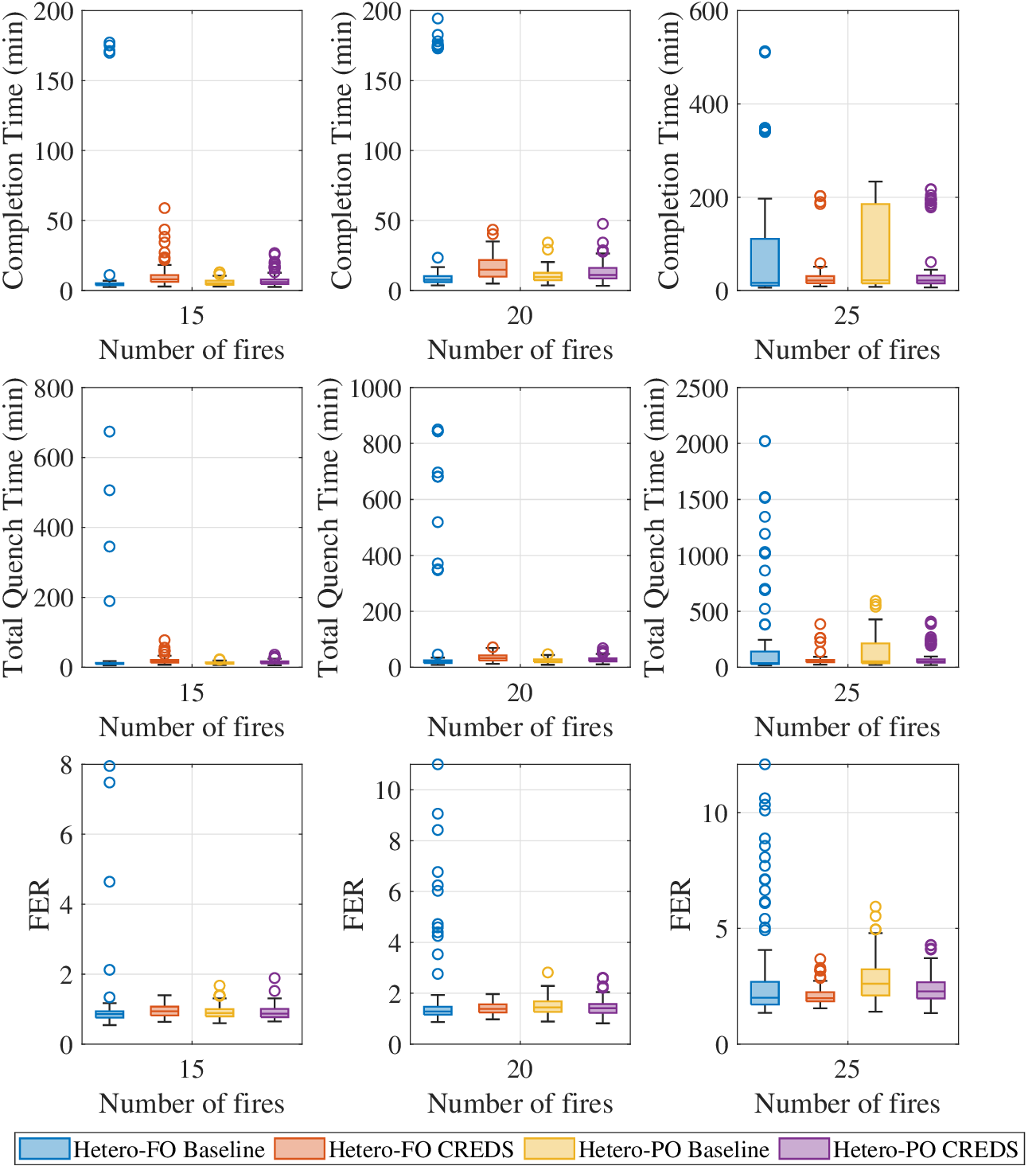}}
	\caption{Box plot of different performance indices for heterogeneous team}
	\label{fig_box_PI_hetero}
\end{figure}

The Monte-Carlo analysis for different numbers of fires using a heterogeneous team of UAVs under partial observability is evaluated. Ablation studies are conducted to examine the impact of homogeneous teams and fully observable conditions. Heterogeneous teams have varying speed and quench capabilities, while homogeneous teams exhibit identical quench and speed capabilities, maintained at the average of the quench area rate and velocity of the heterogeneous team for equitable comparisons. The box plots of performance indices and success rates for different numbers of fires for all analyzed cases are presented in Fig. \ref{fig_box_PI_homo}, Fig. \ref{fig_box_PI_hetero} and Table \ref{tab_SRresults}, respectively. The details of mean values of performance indices are provided in the supplementary material.

\begin{table}[t]
\centering
\caption{Success rate comparison of CREDS with baseline}
\begin{tabular}{|c|c|cccc|}
\hline
\multirow{3}{*}{\begin{tabular}[c]{@{}c@{}}Type of\\ Agents\end{tabular}} & \multirow{3}{*}{$n_t$} & \multicolumn{4}{c|}{Success Rate ($\%$)}                                                           \\ \cline{3-6} 
                                                                          &                        & \multicolumn{2}{c|}{Homogeneous}                           & \multicolumn{2}{c|}{Heterogeneous}    \\ \cline{3-6} 
                                                                          &                        & \multicolumn{1}{c|}{Baseline} & \multicolumn{1}{c|}{CREDS} & \multicolumn{1}{c|}{Baseline} & CREDS \\ \hline
\multirow{3}{*}{FO}                                                       & $15$                   & \multicolumn{1}{c|}{$100$}    & \multicolumn{1}{c|}{$100$} & \multicolumn{1}{c|}{$96$}     & $100$ \\ \cline{2-6} 
                                                                          & $20$                   & \multicolumn{1}{c|}{$93$}     & \multicolumn{1}{c|}{$100$} & \multicolumn{1}{c|}{$89$}     & $100$ \\ \cline{2-6} 
                                                                          & $25$                   & \multicolumn{1}{c|}{$78$}     & \multicolumn{1}{c|}{$91$}  & \multicolumn{1}{c|}{$75$}     & $96$  \\ \hline
\multirow{3}{*}{PO}                                                       & $15$                   & \multicolumn{1}{c|}{$100$}    & \multicolumn{1}{c|}{$100$} & \multicolumn{1}{c|}{$100$}    & $100$ \\ \cline{2-6} 
                                                                          & $20$                   & \multicolumn{1}{c|}{$100$}    & \multicolumn{1}{c|}{$100$} & \multicolumn{1}{c|}{$100$}    & $100$ \\ \cline{2-6} 
                                                                          & $25$                   & \multicolumn{1}{c|}{$66$}     & \multicolumn{1}{c|}{$71$}  & \multicolumn{1}{c|}{$67$}     & $84$  \\ \hline
\end{tabular}
\label{tab_SRresults}
\end{table}

\subsubsection{Homogeneous Team}
\begin{figure}[t]	
    \centerline{\includegraphics[width=8.8cm]{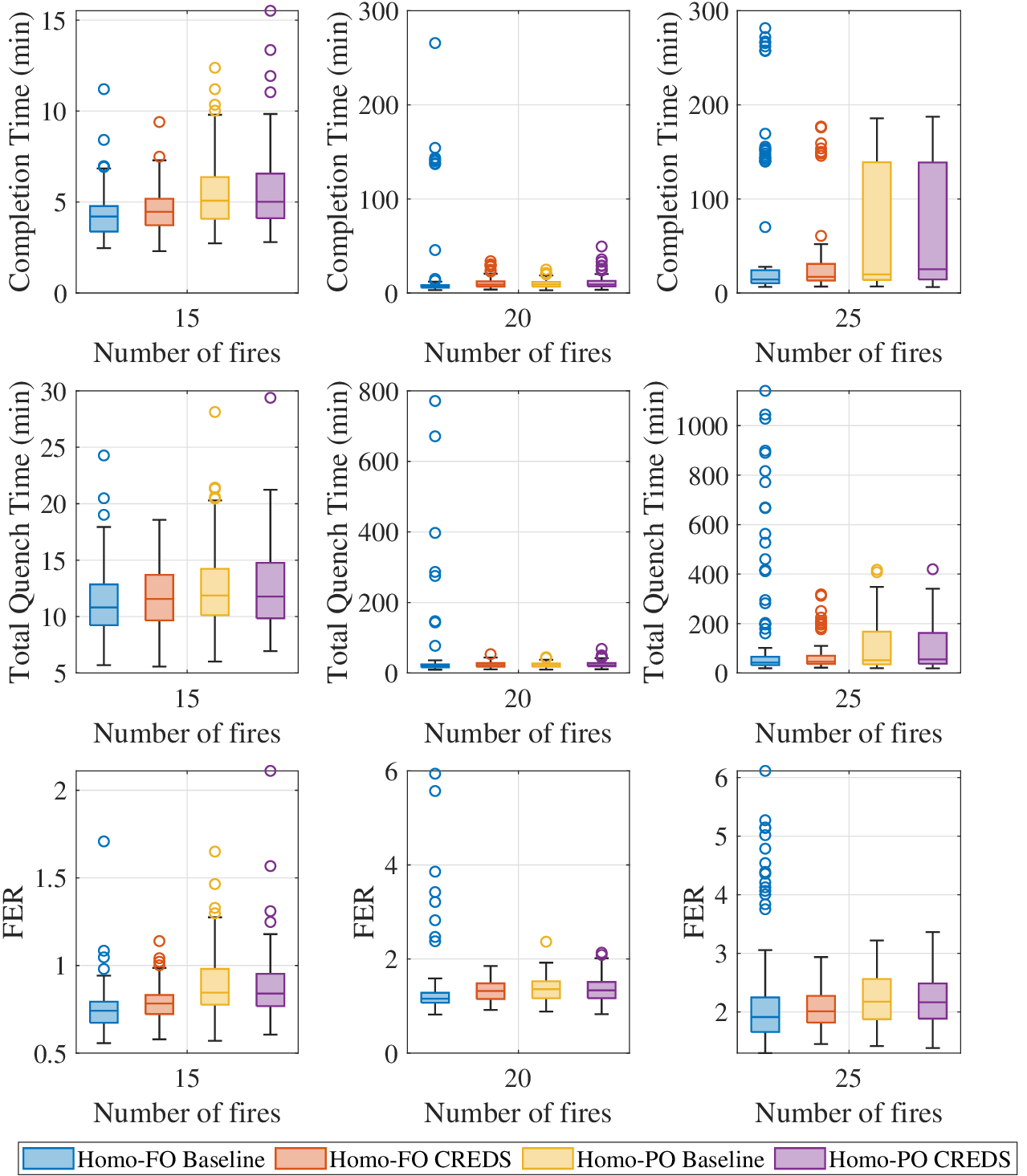}}
	\caption{Box plot of different performance indices for homogeneous team}
	\label{fig_box_PI_homo}
\end{figure}

The performance of a homogeneous team of UAVs with speed, $20 \si{m/s}$, and the quench area rate, $20 \si{m^{2}/s}$, is assessed. The variation performance indices obtained from Monte-Carlo analysis are presented as boxplots in Fig. \ref{fig_box_PI_homo} for homogeneous UAVs with full observability (Homo-FO) and partial observability (Homo-PO) for both baseline and CREDS. The performance indices in Fig. \ref{fig_box_PI_homo} include the cases of failure missions for which the value of performance indices are computed as a very high value. 

The Homo-FO and Homo-PO results indicate that CREDS significantly outperforms the baseline approach with a $100\%$ mission success rate for a fire-to-UAV ratio up to $4$. When mitigating $25$ fires under FO, CREDS achieves a $91\%$ success rate compared to only $78\%$ for the baseline. In the Homo-PO case, as the UAVs are unaware of all the fires, the assignment will be less efficient with more failures than Homo-FO. The values of performance indices will reflect the search time to detect targets, which depends on the effectiveness of the search method. The stochastic search approaches are beneficial in detecting unknown targets in uncertain environments, while the search time associated with stochastic search methods is not constant. Also, the paths generated may not effectively reduce the global reward due to the absence of complete situational awareness. The performance of CREDS is better with a $71\%$ success rate compared to baseline with a $66\%$ success rate for $25$ fires. The performance of the baseline is slightly better than the FO case for $20$ fires as the deadlock conditions are prominent with full observability due to the large number of observed fires. The boxplot of performance indices of FO baseline cases has many outliers compared to other cases.

\subsubsection{Heterogeneous Team}
The performance of the CREDS is assessed for a heterogeneous team with $2$ UAVs having higher quench and speed capabilities and the remaining with low speed and quench rates. The average speed and quench rate of the heterogeneous team are set to match the homogeneous team for a fair comparison. The higher value of speed and quench rate are chosen as $26$, and the lower value as $16$. The variation performance indices obtained from Monte-Carlo analysis are presented as boxplots in Fig. \ref{fig_box_PI_hetero} for heterogeneous UAVs.

The CREDS performs better with a success rate of $96\%$ for $25$ fires and a $100\%$ success rate up to $20$ fires for Hetero-FO. The performance improves compared to the Homo-FO for CREDS in terms of success rate but other performance indices are higher than the homogeneous case. Heterogeneous cases have UAVs with high and low quench capabilities handling heterogeneous deadlines. The deadline time is a function of the quench rate, spread rate, and initial area of the fire. The UAVs with high quench capability increase the deadline time for a given spread rate and fire area. The high-speed, high quench rate UAVs have later deadlines and can mitigate fire areas grown into larger areas, increasing feasible assignments. The algorithm will push the task start times closer to their deadline times to generate feasible paths with more tasks. Thus heterogeneous cases possess the higher advantage of scheduling larger fires in the path compared to homogenous cases. The Hetero-FO baseline case performs worse than the Homo-FO and Homo-PO cases, with increased failures for $15$ and $20$ fires. The major reason for the degradation of baseline performance is the increased probability of exclusion of larger fires with higher observations. Also, the number of deadlock scenarios increases with heterogeneous teams compared to homogeneous teams.

The Hetero-PO results demonstrate a significant advantage for CREDS, achieving a $100\%$ success rate for up to $20$ fires and an $84\%$ success rate for $25$ fires. The success rate is reduced compared to the Hetero-FO case, but the other performance indices have lower values due to the lower number of observations, number of tasks pushed to deadlines becomes less compared to the Hetero-FO case. The partial observability makes it harder for CREDS to optimize the paths as the number of fires increases. With more fires, the trend reverses, reflecting an increase in mean completion time, quench time, and FER compared to Hetero-FO. This highlights that the challenges of partial observability become more critical when dealing with larger wildfire scenarios. The baseline has similar performance up to $20$ fires and deteriorates considerably in the case of $25$ fires.The CREDS improves the success rate by $21\%$ for Hetero-FO and $17\%$ for Hetero-PO compared to baseline.

\subsubsection{Scalability}
\begin{figure}[t]
	\begin{minipage}{.5\linewidth}
		\centering
		\subfloat[]{\label{fig_QR:a}\includegraphics[trim=4 1 .5 1,clip,width=4.35cm]{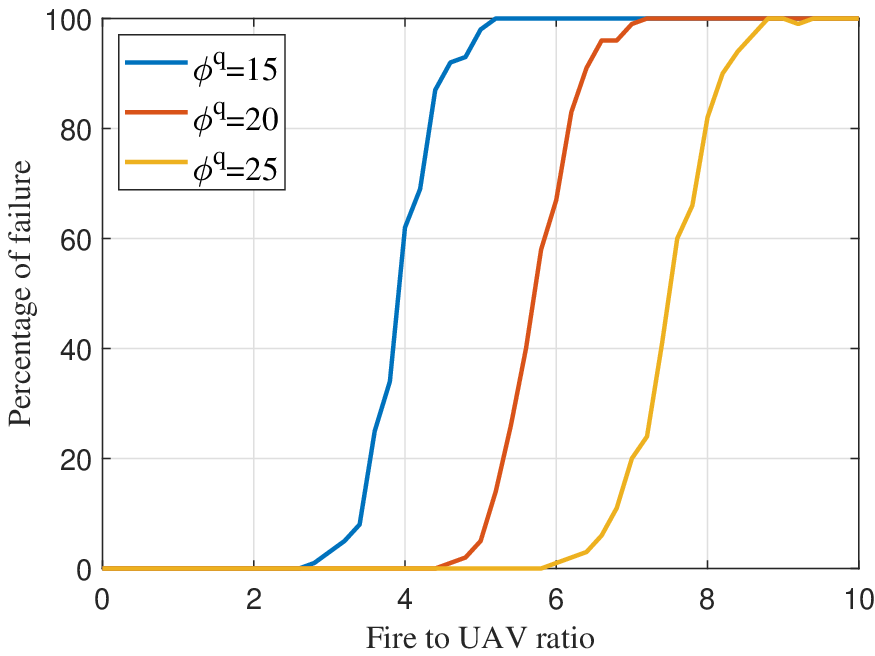}}
	\end{minipage}%
	\begin{minipage}{.5\linewidth}
		\centering
		\subfloat[]{\label{fig_VEL:b}\includegraphics[trim=4 1 .5 1,clip,width=4.35cm]{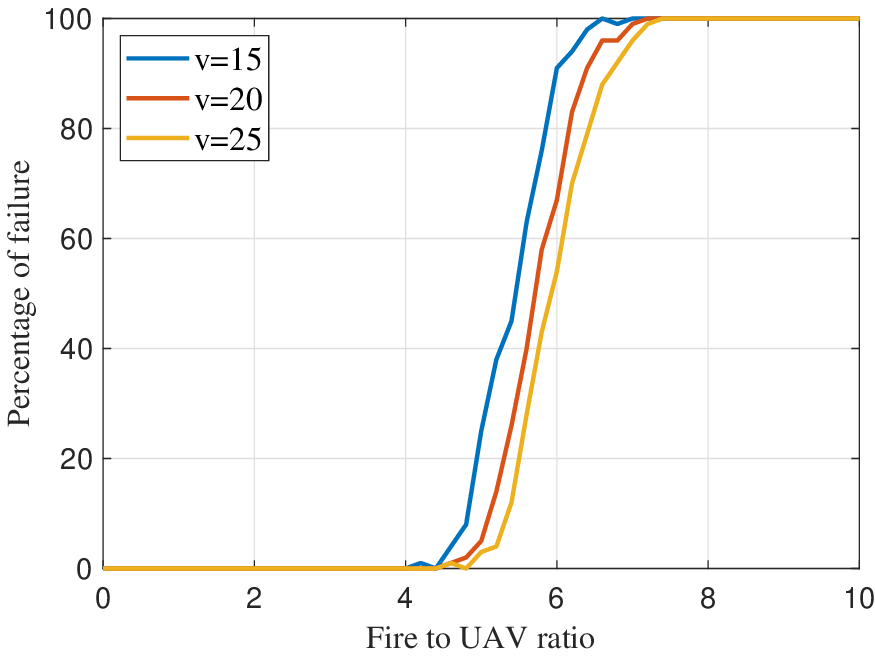}}
	\end{minipage}
	\caption{Percentage of failure vs fire to UAV ratio (a) Variation of quench rate for constant velocity. (b)Variation of velocity for constant quench rate.}
	\label{fig_ratio_failure_diff_QRVEL}	
\end{figure}

The percentage failure for different fire-to-UAV ratios is plotted in Fig. \ref{fig_ratio_failure_diff_QRVEL} with different quench rates and velocities of a homogeneous team. The effect of quench rate variation with a constant velocity of $20$ \si{m/s} is shown in Fig. \ref{fig_QR:a} and the effect of velocity variation with a constant quench rate of $20$ \si{m^2/s} is shown in \ref{fig_VEL:b}. The increase in quench rate improves the success rates considerably for higher fire-to-UAV ratios. The percentage of failures steadily increases beyond the threshold ratio corresponding to each quench rate. However, the increase in the success rates for higher fire-to-UAV ratios with the increase in velocity is less. Thus for a given wildfire management scenario with specified fire spread rate, the quench rate plays a critical role compared to the velocity. Thus, the quench rate, velocity and number of UAVs should be carefully chosen to build a heterogeneous team to achieve a successful mission for scalable wildfire scenarios.
\begin{table}[t]
\centering
\caption{Convergence rate and average number of iterations for convergence}
\setlength{\tabcolsep}{3.6pt}
\begin{tabular}{|c|cccc|cccc|}
\hline
\multirow{3}{*}{$n_t$} & \multicolumn{4}{c|}{Homogeneous}                                                                           & \multicolumn{4}{c|}{Heterogeneous}                                                                         \\ \cline{2-9} 
                       & \multicolumn{2}{c|}{Baseline}                                  & \multicolumn{2}{c|}{CREDS}                & \multicolumn{2}{c|}{Baseline}                                  & \multicolumn{2}{c|}{CREDS}                \\ \cline{2-9} 
                       & \multicolumn{1}{c|}{CR($\%$)} & \multicolumn{1}{c|}{$I_{avg}$} & \multicolumn{1}{c|}{CR($\%$)} & $I_{avg}$ & \multicolumn{1}{c|}{CR($\%$)} & \multicolumn{1}{c|}{$I_{avg}$} & \multicolumn{1}{c|}{CR($\%$)} & $I_{avg}$ \\ \hline
$15$                   & \multicolumn{1}{c|}{$99$}     & \multicolumn{1}{c|}{$9.02$}    & \multicolumn{1}{c|}{$100$}    & $7.36$    & \multicolumn{1}{c|}{$94$}     & \multicolumn{1}{c|}{$10.07$}   & \multicolumn{1}{c|}{$100$}    & $7.36$    \\ \hline
$20$                   & \multicolumn{1}{c|}{$91$}     & \multicolumn{1}{c|}{$10.90$}   & \multicolumn{1}{c|}{$100$}    & $8.89$    & \multicolumn{1}{c|}{$87$}     & \multicolumn{1}{c|}{$11.77$}   & \multicolumn{1}{c|}{$100$}    & $8.95$    \\ \hline
$25$                   & \multicolumn{1}{c|}{$83$}     & \multicolumn{1}{c|}{$12.75$}   & \multicolumn{1}{c|}{$100$}    & $10.33$   & \multicolumn{1}{c|}{$78$}     & \multicolumn{1}{c|}{$13.65$}   & \multicolumn{1}{c|}{$100$}    & $10.80$   \\ \hline
\end{tabular}
\label{tab_convergence}
\end{table}
\subsubsection{Convergence}
In wildfire management, the cost function exhibits non-stationary behavior, leading to deadlock conditions. The conflict-free assignment toggles at each iteration, preventing algorithm convergence. This deadlock phenomenon is particularly pronounced in baseline cases, prompting the introduction of a deadlock removal mechanism in the consensus algorithm to finalize assignments. The deadlock removal mechanism enhances the success rate of the baseline by selecting assignments with the least infeasible tasks. Convergence rates (CR) for both baseline and CREDS are analyzed under full observability conditions, with results shown in Table \ref{tab_convergence}. Baseline cases often fail to achieve convergence even with a smaller number of fires, and non-convergence or deadlock scenarios are increased in heterogeneous baseline scenarios.

The comparison of the average number of iterations ($I_{avg}$) required for convergence is also provided in Table \ref{tab_convergence} for each case. The CREDS with DPMC is more efficient and robust to achieve a $100\%$ convergence with a lesser number of average iterations for all the case studies without deadlock removal assistance. The maximum limit of iterations is fixed at $3m$ for the simulation studies where $m$ is the number of UAVs. The increase in the maximum limit beyond $3m$ exhibited negligible impact on the success rate and convergence behavior of the baseline cases. The baseline approach consistently required at least $20\%$ more iterations compared to CREDS, resulting in significantly higher computational demands. In the $25$ fire scenario with a heterogeneous team, CREDS achieves a $100\%$ convergence rate with $26.3\%$ fewer iterations, while the baseline approach exhibits lower convergence at $78\%$, with deadlocks occurring in $22\%$ of Monte-Carlo simulations. This translates to a significant reduction in computation time, making CREDS a more scalable solution for real-world wildfire management scenarios where timely intervention is important.

\section{Conclusions}
This paper presents Conflict-aware Resource-Efficient Decentralized Sequential planner for early wildfire mitigation using multiple heterogeneous UAVs. CREDS addresses the challenges posed by non-stationary wildfire scenarios with dynamically spreading fires, potential pop-up fires, and partial observability due to limited UAV numbers and sensing range. CREDS focuses on timely detection and sequential mitigation of growing fires as SUTs, thereby preventing the escalation of fires into MUTs and minimizing overall loss of biodiversity with efficient resource utilization. CREDS reformulates resource-limited wildfire management as a decentralized sequential spatiotemporal task assignment for growing tasks. The CREDS employs a three-phased framework, including OMS for the search phase, REDS for local trajectory generation, and conflict-aware consensus algorithm for global trajectory generation. A novel non-stationary deadline-prioritized mitigation cost is proposed to efficiently prioritize tasks with heterogeneous deadlines to achieve high success and convergence rates. The performance of CREDS is evaluated using Monte-Carlo analysis for partial and full observability conditions with both heterogeneous and homogeneous UAV teams for different fires-to-UAV ratios. CREDS demonstrates superior performance in Monte-Carlo analysis compared to the baseline approach. In critical scenarios with the fire-to-UAV ratio of $5$, CREDS achieves a $17\%$ higher success rate for partially observable heterogeneous teams compared to the baseline. Moreover, the heterogeneous team obtains a $13\%$ higher success rate than homogeneous teams. CREDS further demonstrates efficiency with full observability case, attaining a $21\%$ improvement in success rate with $100\%$ convergence in $26.3\%$ fewer iterations. CREDS exhibits high success rates, robustness against deadlock scenarios, and scalability in resource-limited environments, making it suitable for real-world applications with growing costs and time constraints.
 
\bibliographystyle{IEEEtran}
\balance
\bibliography{forestfire_SRT}

\pagebreak
\nobalance
\setcounter{page}{1}
\section*{Supplementary Material}

\subsection{Working of CREDS}
 \begin{figure*}[b]
		\centering
		\subfloat[] {\label{fig_sim:a}
		\includegraphics[trim=30 2 65 2,clip,width=4.4cm]{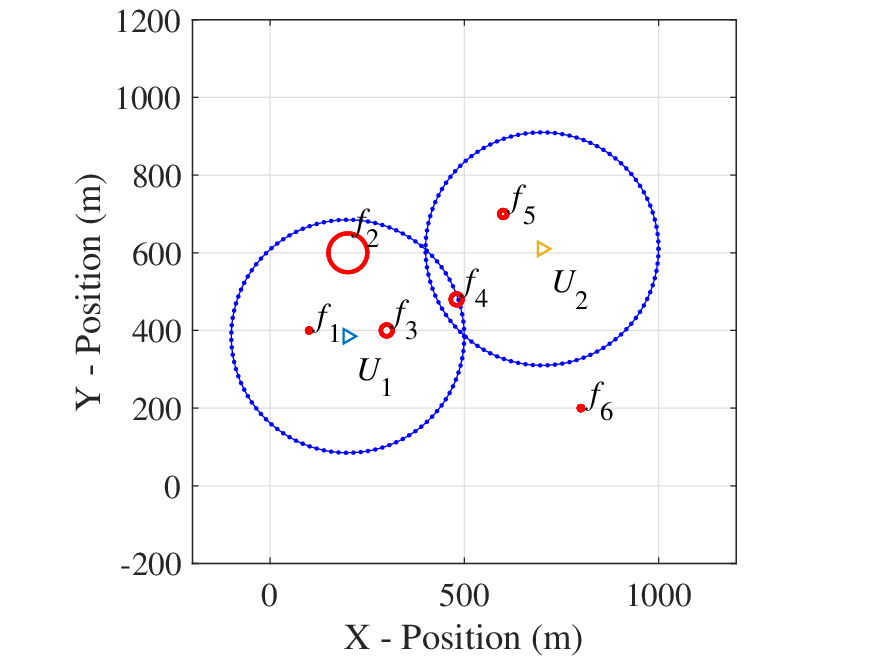}}%
		\subfloat[]{\label{fig_sim:b}
			\includegraphics[trim=30 2 65 2,clip,width=4.4cm]{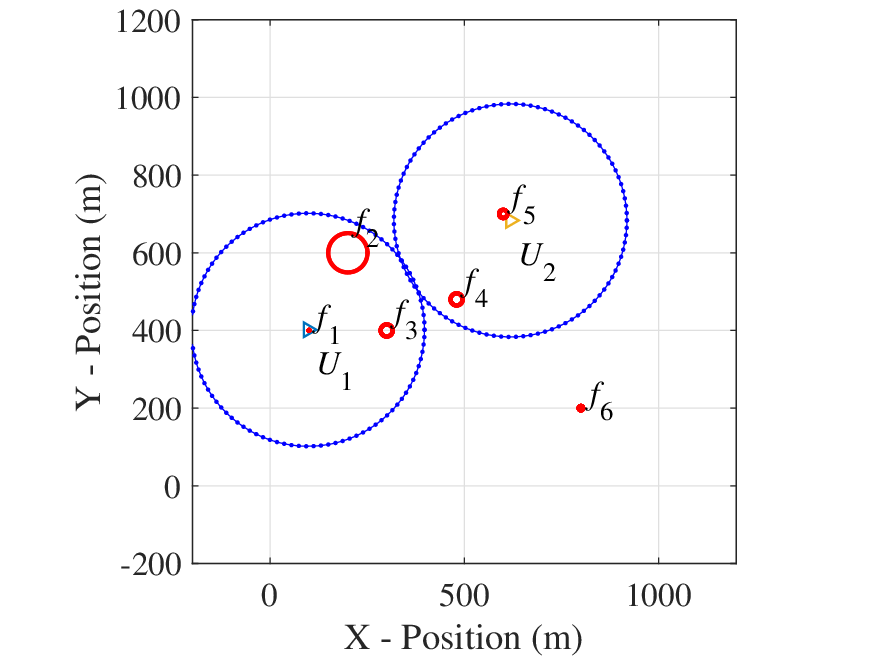}}%
		\subfloat[]{\label{fig_sim:c}
			\includegraphics[trim=30 2 65 2,clip,width=4.4cm]{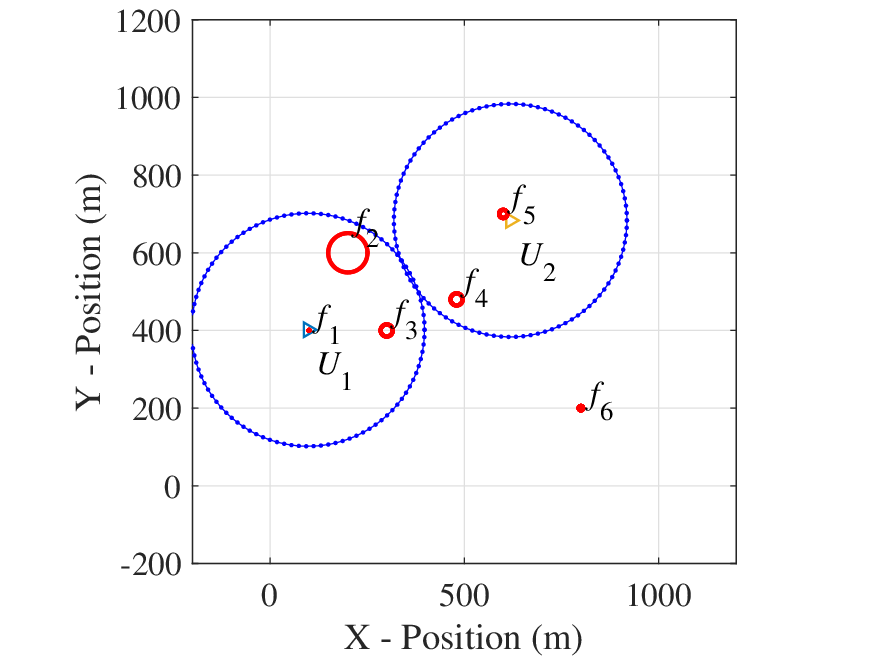}}%
		\subfloat[]{\label{fig_sim:d}
			\includegraphics[trim=30 2 65 2,clip,width=4.4cm]{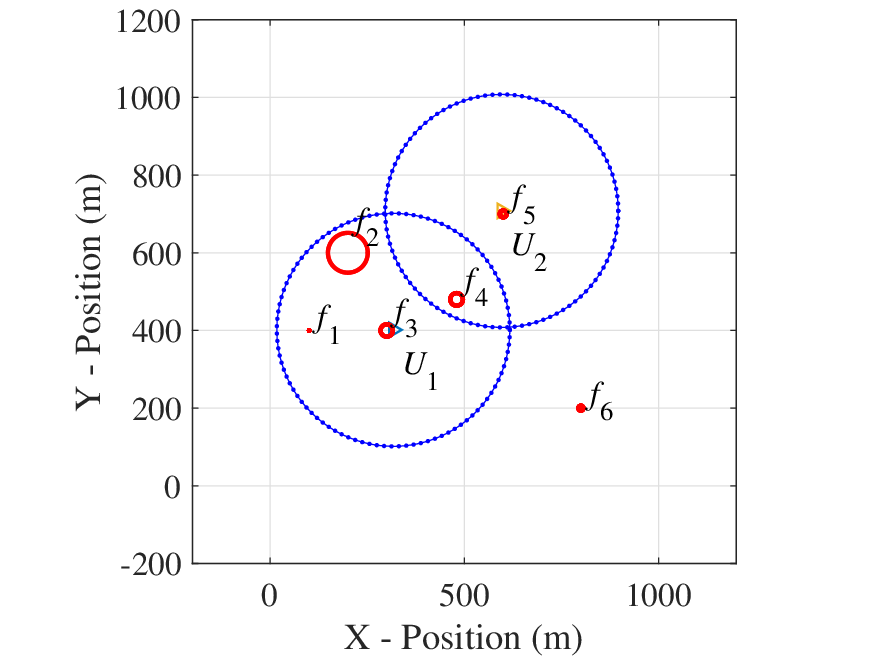}}\\
            \subfloat[]{\label{fig_sim:e}
			\includegraphics[trim=30 2 65 2,clip,width=4.4cm]{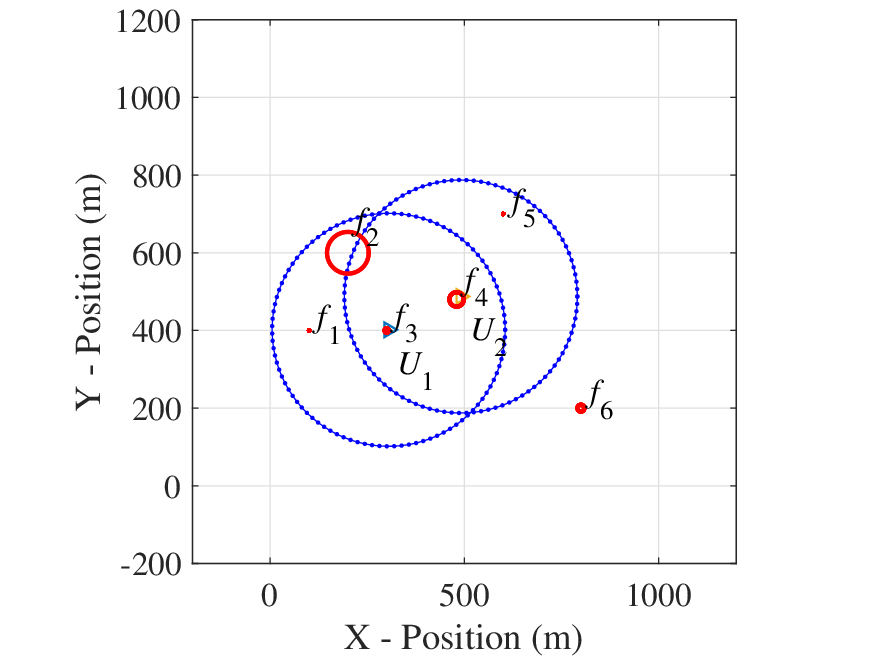}}
            \subfloat[]{\label{fig_sim:f}
			\includegraphics[trim=30 2 65 2,clip,width=4.4cm]{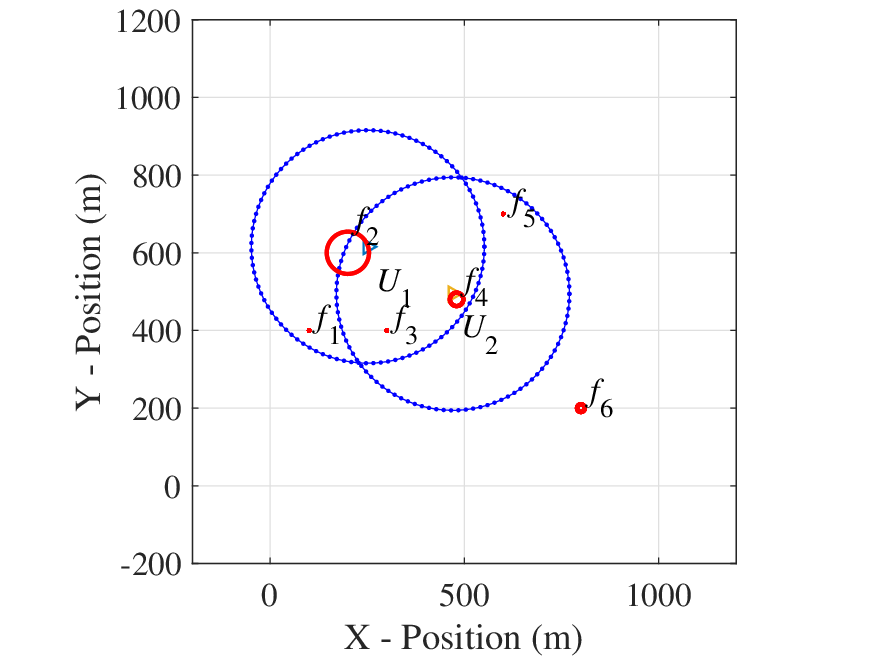}}
             \subfloat[]{\label{fig_sim:g}
			\includegraphics[trim=30 2 65 2,clip,width=4.4cm]{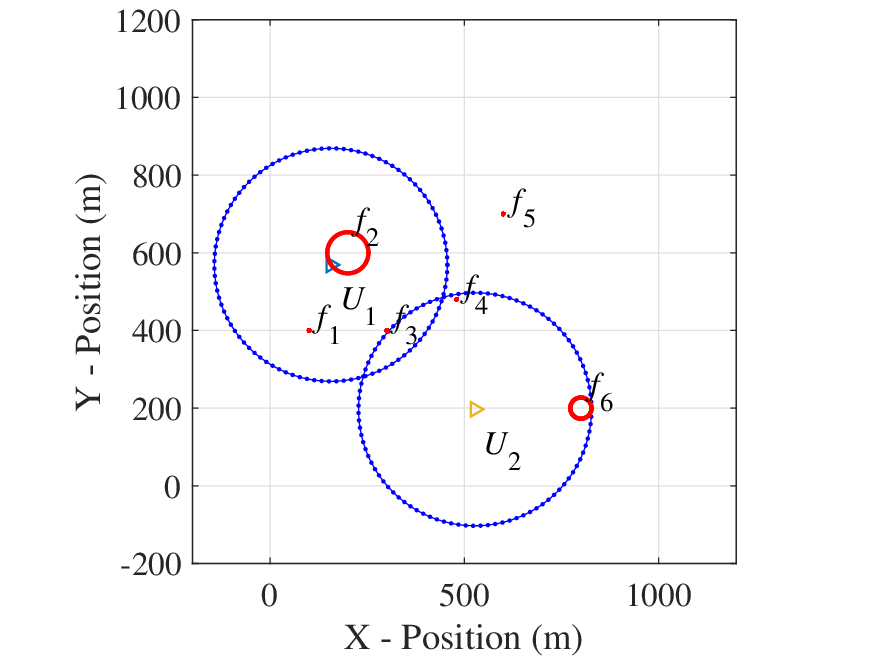}}
              \subfloat[]{\label{fig_sim:h}
			\includegraphics[trim=30 2 65 2,clip,width=4.4cm]{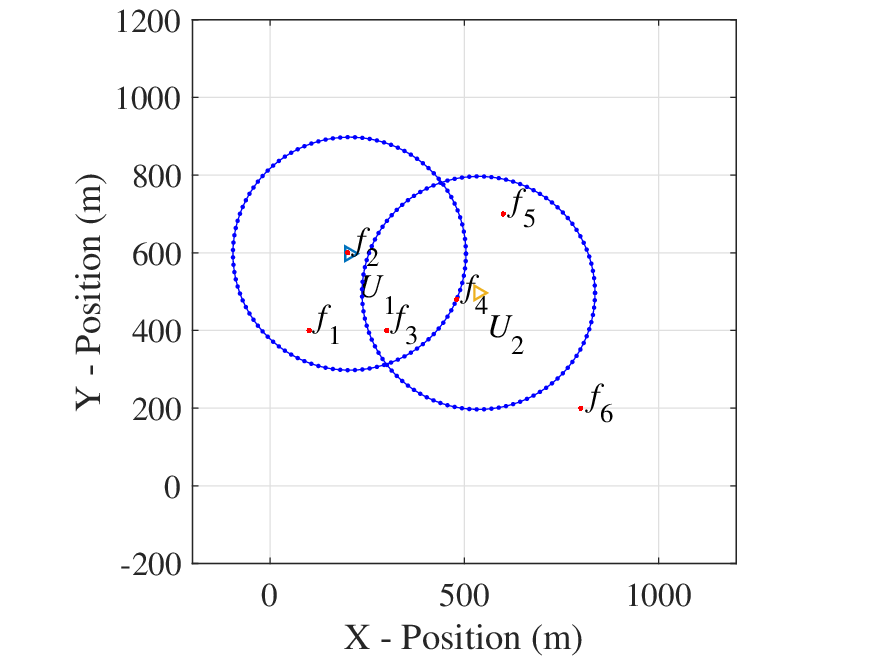}}
		\caption{Working of CREDS: (a) Initial positions of UAVs and fire locations  (b) UAV, $U_1$ is assigned for fire, $f_1$, (c) UAV, $U_2$ is assigned for fire, $f_5$ (d) After quenching $f_1$, $U_1$ moves to $f_3$ in its path, (e) $U_2$ is assigned for fire, $f_4$ (f) After quenching $f_3$, $U_1$ moves to the large fire area $f_1$ in its path,(g) $U_2$ with no observations use OMS to search for fire locations and $U_2$ senses fire $f_5$ (h) UAVs finish tasks in their path and continue searching the mission area for new fires.}
		\label{fig_CREDS_plannersteps}
\end{figure*}
 A firefighting scenario with six fire locations and two UAVs is considered as a demonstrative case to explain the working of CREDS. The center coordinates of the six fire locations are $(100, 400)$ \si{m}, $(200, 600)$ \si{m}, $(300, 400)$ \si{m}, $(480, 480)$ \si{m}, $(600, 700)$ \si{m}, $(800, 200)$ \si{m}. The initial fire radii are $5$ \si{m}, $50$ \si{m}, $15$ \si{m}, $15$ \si{m}, $10$ \si{m}, and $5$ \si{m} respectively. Let the initial positions of two UAVs be $(200, 385)$ \si{m} and $(700, 610)$ \si{m}. The $U_1$ has higher speed and quench capabilities than $U_2$. The UAVs have partial observability and sense the tasks located within their sensing radius of $300$ \si{m}. The different steps associated with the working of CREDS are shown in  Fig. \ref{fig_CREDS_plannersteps}. The UAV, $U_1$, can observe fires, $f_1$, $f_2$, $f_3$, and $f_4$ and UAV, $U_2$, can observe fires $f_4$ and $f_5$ as shown in Fig. \ref{fig_sim:a}. The UAVs compute their bids for all observed tasks from their initial position and identify their valid tasks. A task is valid for a UAV if its bid is smaller than the known winning bid value. The task having the smallest bid from the valid task list gets added to the bundles and paths of UAVs. The tasks are added to the location of the path to obtain the minimum score improvement. The bundle generated using DPMC for the $U_1$ and $U_2$ from their initial position is $[2, 1, 3]$ and$[5, 4]$ and their corresponding paths are $[1, 3, 2]$ and $[5, 4]$. The largest fire with an earlier deadline gets added to the bundle of $U_1$ first, and tasks are strategically positioned to achieve a feasible path. The bid, $C_{14}$ becomes $\infty$ as $f_4$ is infeasible due to the large fire area in the path. The bid, $C_{24}$ of $U_2$ for $f_4$ is the lowest bid and $U_2$ wins $f_4$ during conflict resolution. The UAVs move to the fire location and quench the fire by spraying water along the firefront. After finishing the assigned tasks, the UAVs move to the next task in the path. The UAVs replan their path if there is any new observation in their detected task list. The different time frames of UAVs following the generated path for fire mitigation are shown in Fig. \ref{fig_sim:a}- \ref{fig_sim:f}. The $U_2$ employs OMS search after the mitigation of $f_4$ as there are no detected fires. The OMS helps in the exploration and exploitation of the search space, and Fig. \ref{fig_sim:g} shows the detection of $f_6$ by $U_2$ before the deadline time. After the mitigation of $f_1$ and $f_6$, both UAVs $U_1$ and $U_2$ have zero fire detections, and Fig. \ref{fig_sim:h} shows UAVs employing OMS for detecting possible fire areas. The scenario is also analyzed for a full observability scenario where all the fires are observable to all UAVs. The feasible paths generated using CREDS with DPMC is $\mu_1=[1, 3, 2]$ and $\mu_2=[5, 6, 4]$.
 
 The same scenario is analyzed with baseline cost, and the path generated was $[1, 3, 4]$ and $[5, 6]$ for partial observability. The baseline fails to generate a feasible path due to the exclusion $f_2$. The baseline adds tasks with the smallest completion time, and $f_2$ is not added to the bundle in the initial iterations. The baseline fails to generate a feasible path for this scenario, even for the full observability scenario. The path generated for full observability scenario is $[1, 3, 6]$ and $[5, 4]$.

\subsection{Monte-Carlo Analysis- Mean Performance Indices}

The mean value of performance indices for CREDS and baseline for homogeneous and heterogeneous teams are tabulated in Table \ref{tab_HOMOresults} and \ref{tab_HETEROresults}, respectively. The performance indices value with $*$ shown in Table \ref{tab_HOMOresults} and \ref{tab_HETEROresults} indicates the failure missions for which the values of performance indices are computed as a very high value. The actual value of performance indices for failure cases will be $\infty$ for a single UAV. In smaller fire scenarios, where both the baseline and CREDS achieve a $100\%$ success rate, CREDS exhibits slightly higher mean completion time and mean total quench time compared to the baseline. The CREDS prioritizes tasks with earlier deadlines and tends to accomplish tasks near deadlines, increasing the mean performance indices value.  But for failure cases, the baseline has a higher mean completion time and mean total quench time. 

\begin{table*}[t]
\centering
\caption{Performance comparison of CREDS with baseline case for homogeneous team}
\begin{tabular}{cccccccccc}
\hline
\multicolumn{1}{|c|}{\multirow{2}{*}{\begin{tabular}[c]{@{}c@{}}Type of\\ Agents\end{tabular}}} & \multicolumn{1}{c|}{\multirow{2}{*}{$n_t$}} & \multicolumn{2}{c|}{Success rate(\%)}                      & \multicolumn{2}{c|}{Mean Completion Time (min)}                           & \multicolumn{2}{c|}{Mean Total Quench Time (min)}                         & \multicolumn{2}{c|}{Mean FER}                                           \\ \cline{3-10} 
\multicolumn{1}{|c|}{}                                                                          & \multicolumn{1}{c|}{}                       & \multicolumn{1}{c|}{Baseline} & \multicolumn{1}{c|}{CREDS} & \multicolumn{1}{c|}{Baseline}       & \multicolumn{1}{c|}{CREDS}          & \multicolumn{1}{c|}{Baseline}       & \multicolumn{1}{c|}{CREDS}          & \multicolumn{1}{c|}{Baseline}      & \multicolumn{1}{c|}{CREDS}         \\ \hline
\multicolumn{1}{|c|}{\multirow{3}{*}{FO}}                                                       & \multicolumn{1}{c|}{$15$}                   & \multicolumn{1}{c|}{$100$}    & \multicolumn{1}{c|}{$100$} & \multicolumn{1}{c|}{$4.29\pm1.32$}  & \multicolumn{1}{c|}{$4.55\pm1.23$}  & \multicolumn{1}{c|}{$11.24\pm3.07$} & \multicolumn{1}{c|}{$11.64\pm2.76$} & \multicolumn{1}{c|}{$0.75\pm0.13$} & \multicolumn{1}{c|}{$0.79\pm0.10$} \\ \cline{2-10} 
\multicolumn{1}{|c|}{}                                                                          & \multicolumn{1}{c|}{$20$}                   & \multicolumn{1}{c|}{$93$}     & \multicolumn{1}{c|}{$100$} & \multicolumn{1}{c|}{$18.12*$}       & \multicolumn{1}{c|}{$10.67\pm6.18$} & \multicolumn{1}{c|}{$46.02*$}       & \multicolumn{1}{c|}{$23.98\pm8.61$} & \multicolumn{1}{c|}{$1.35*$}       & \multicolumn{1}{c|}{$1.32\pm0.22$} \\ \cline{2-10} 
\multicolumn{1}{|c|}{}                                                                          & \multicolumn{1}{c|}{$25$}                   & \multicolumn{1}{c|}{$78$}     & \multicolumn{1}{c|}{$91$}  & \multicolumn{1}{c|}{$51.96*$}       & \multicolumn{1}{c|}{$32.97*$}       & \multicolumn{1}{c|}{$153.00*$}      & \multicolumn{1}{c|}{$64.48*$}       & \multicolumn{1}{c|}{$2.26*$}       & \multicolumn{1}{c|}{$2.06*$}       \\ \hline
\multicolumn{1}{|c|}{\multirow{3}{*}{PO}}                                                       & \multicolumn{1}{c|}{$15$}                   & \multicolumn{1}{c|}{$100$}    & \multicolumn{1}{c|}{$100$} & \multicolumn{1}{c|}{$5.43\pm1.97 $} & \multicolumn{1}{c|}{$5.61\pm2.18$}  & \multicolumn{1}{c|}{$12.44\pm3.66$} & \multicolumn{1}{c|}{$12.45\pm3.65$} & \multicolumn{1}{c|}{$0.89\pm0.19$} & \multicolumn{1}{c|}{$0.88\pm0.20$} \\ \cline{2-10} 
\multicolumn{1}{|c|}{}                                                                          & \multicolumn{1}{c|}{$20$}                   & \multicolumn{1}{c|}{$100$}    & \multicolumn{1}{c|}{$100$} & \multicolumn{1}{c|}{$9.80\pm4.17$}  & \multicolumn{1}{c|}{$11.20\pm7.27$} & \multicolumn{1}{c|}{$23.13\pm7.45$} & \multicolumn{1}{c|}{$24.46\pm9.63$} & \multicolumn{1}{c|}{$1.36\pm0.25$} & \multicolumn{1}{c|}{$1.37\pm0.26$} \\ \cline{2-10} 
\multicolumn{1}{|c|}{}                                                                          & \multicolumn{1}{c|}{$25$}                   & \multicolumn{1}{c|}{$66$}     & \multicolumn{1}{c|}{$71$}  & \multicolumn{1}{c|}{$60.79*$}       & \multicolumn{1}{c|}{$59.35*$}       & \multicolumn{1}{c|}{$101.90*$}      & \multicolumn{1}{c|}{$100.04*$}      & \multicolumn{1}{c|}{$2.21*$}       & \multicolumn{1}{c|}{$2.19*$}       \\ \hline
\multicolumn{10}{l}{\scriptsize{$*$ indicates failure cases}}                                                                                                                                                                                                                                                                                                                                                                                              
\end{tabular}
\label{tab_HOMOresults}
\end{table*}

\begin{table*}[t]
\centering
\caption{Performance comparison of CREDS with baseline case for heterogeneous team}
\begin{tabular}{cccccccccc}
\hline
\multicolumn{1}{|c|}{\multirow{2}{*}{\begin{tabular}[c]{@{}c@{}}Type of\\ Agents\end{tabular}}} & \multicolumn{1}{c|}{\multirow{2}{*}{$n_t$}} & \multicolumn{2}{c|}{Success rate ($\%$)}                   & \multicolumn{2}{c|}{Mean Completion Time (min)}                           & \multicolumn{2}{c|}{Mean Total Quench Time (min)}                           & \multicolumn{2}{c|}{Mean FER}                                           \\ \cline{3-10} 
\multicolumn{1}{|c|}{}                                                                          & \multicolumn{1}{c|}{}                       & \multicolumn{1}{c|}{Baseline} & \multicolumn{1}{c|}{CREDS} & \multicolumn{1}{c|}{Baseline}       & \multicolumn{1}{c|}{CREDS}          & \multicolumn{1}{c|}{Baseline}       & \multicolumn{1}{c|}{CREDS}            & \multicolumn{1}{c|}{Baseline}      & \multicolumn{1}{c|}{CREDS}         \\ \hline
\multicolumn{1}{|c|}{\multirow{3}{*}{FO}}                                                       & \multicolumn{1}{c|}{$15$}                   & \multicolumn{1}{c|}{$96$}     & \multicolumn{1}{c|}{$100$} & \multicolumn{1}{c|}{$11.32*$}       & \multicolumn{1}{c|}{$10.39\pm8.54$} & \multicolumn{1}{c|}{$27.44*$}       & \multicolumn{1}{c|}{$19.22\pm10.77$}  & \multicolumn{1}{c|}{$1.03*$}       & \multicolumn{1}{c|}{$0.95\pm0.17$} \\ \cline{2-10} 
\multicolumn{1}{|c|}{}                                                                          & \multicolumn{1}{c|}{$20$}                   & \multicolumn{1}{c|}{$89$}     & \multicolumn{1}{c|}{$100$} & \multicolumn{1}{c|}{$26.46*$}       & \multicolumn{1}{c|}{$16.87\pm8.64$} & \multicolumn{1}{c|}{$82.71*$}       & \multicolumn{1}{c|}{$35.43\pm 13.81$} & \multicolumn{1}{c|}{$1.82*$}       & \multicolumn{1}{c|}{$1.40\pm0.20$} \\ \cline{2-10} 
\multicolumn{1}{|c|}{}                                                                          & \multicolumn{1}{c|}{$25$}                   & \multicolumn{1}{c|}{$75$}     & \multicolumn{1}{c|}{$96$}  & \multicolumn{1}{c|}{$76.17*$}       & \multicolumn{1}{c|}{$30.25*$}       & \multicolumn{1}{c|}{$233.15*$}      & \multicolumn{1}{c|}{$65.29*$}         & \multicolumn{1}{c|}{$2.95*$}       & \multicolumn{1}{c|}{$2.08*$}       \\ \hline
\multicolumn{1}{|c|}{\multirow{3}{*}{PO}}                                                       & \multicolumn{1}{c|}{$15$}                   & \multicolumn{1}{c|}{$100$}    & \multicolumn{1}{c|}{$100$} & \multicolumn{1}{c|}{$5.67\pm2.11$}  & \multicolumn{1}{c|}{$7.45\pm4.71$}  & \multicolumn{1}{c|}{$12.43\pm3.42$} & \multicolumn{1}{c|}{$14.90\pm5.77$}   & \multicolumn{1}{c|}{$0.92\pm0.19$} & \multicolumn{1}{c|}{$0.91\pm0.19$} \\ \cline{2-10} 
\multicolumn{1}{|c|}{}                                                                          & \multicolumn{1}{c|}{$20$}                   & \multicolumn{1}{c|}{$100$}    & \multicolumn{1}{c|}{$100$} & \multicolumn{1}{c|}{$10.42\pm4.77$} & \multicolumn{1}{c|}{$12.95\pm7.00$} & \multicolumn{1}{c|}{$23.73\pm7.87$} & \multicolumn{1}{c|}{$28.04\pm10.64$}  & \multicolumn{1}{c|}{$1.49\pm0.32$} & \multicolumn{1}{c|}{$1.45\pm0.33$} \\ \cline{2-10} 
\multicolumn{1}{|c|}{}                                                                          & \multicolumn{1}{c|}{$25$}                   & \multicolumn{1}{c|}{$67$}     & \multicolumn{1}{c|}{$84$}  & \multicolumn{1}{c|}{$76.10*$}       & \multicolumn{1}{c|}{$48.67*$}       & \multicolumn{1}{c|}{$125.14*$}      & \multicolumn{1}{c|}{$84.03*$}         & \multicolumn{1}{c|}{$2.77*$}       & \multicolumn{1}{c|}{$2.38*$}       \\ \hline
\multicolumn{10}{l}{\scriptsize{$*$ indicates failure cases}}                   \end{tabular}
\label{tab_HETEROresults}
\end{table*}

\begin{figure*}
	\centerline{\includegraphics[trim=58 15 68 16,clip,width=14cm]{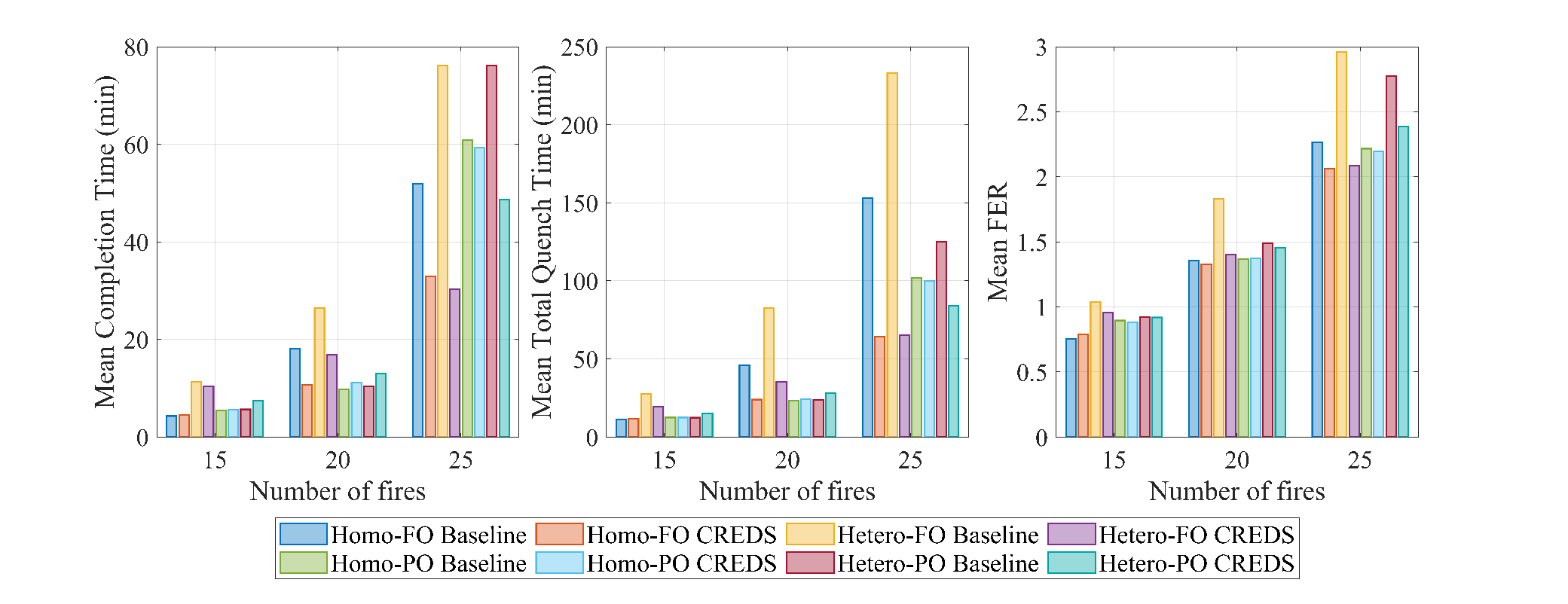}}
	\caption{Bar plot of mean performance indices.}
	\label{fig_bar_FER}
\end{figure*}

The mean performance indices of the heterogeneous team are higher compared to the homogeneous case. The deadline time is a function of the quench rate, spread rate, and initial area of the fire. Heterogeneous cases have teams of UAVs with higher and lower quench capabilities than homogeneous teams. The UAVs with high quench capability increase the deadline time for a given spread rate and fire area. The high-speed, high quench rate UAVs have later deadlines and can mitigate fire areas grown into larger areas, increasing feasible assignments. The UAVs with low quench capability and low speed have smaller deadlines, but the quench time with low-capacity UAVs will be higher, increasing the quench and completion times. The heterogeneous team handles the heterogeneous deadlines more efficiently with a higher success rate compared to the homogeneous team. Even though the completion time and quench time of CREDS are higher than the baseline for certain cases, CREDS has a lower or equal mean FER for all heterogeneous cases, showing reduced loss of biodiversity compared to the baseline. The quench time increases non-linearly with the fire area, particularly when the area is close to the critical area. This phenomenon can lead to a scenario where the total quench time for multiple fires is significant, even if the total fire area is relatively modest. Furthermore, low quench rate UAVs may also be assigned to extinguish fires nearing criticality, resulting in extended quench times despite a potentially smaller overall fire area increase along their designated path.

\end{document}